\documentclass{article}

    \usepackage[preprint]{neurips_2024}

\usepackage{amsmath,amsfonts,bm}

\def\eqref#1{equation~\ref{#1}}

\def\1{\bm{1}}

\DeclareMathAlphabet{\mathsfit}{\encodingdefault}{\sfdefault}{m}{sl}
\SetMathAlphabet{\mathsfit}{bold}{\encodingdefault}{\sfdefault}{bx}{n}

\usepackage{tikz}
\usetikzlibrary{shapes.geometric, arrows}
\usetikzlibrary{backgrounds}
\usepackage{macros}
\usepackage{lmodern}
\usepackage[most]{tcolorbox}
\usepackage{amsmath}
\usepackage{peanutgallery}
\usepackage{xcolor, listings}
\usepackage{textcomp}
\usepackage{enumitem}
\usepackage[utf8]{inputenc} 
\usepackage[T1]{fontenc}    
\usepackage{hyperref}       
\usepackage{url}            
\usepackage{booktabs}       
\usepackage{amsfonts}       
\usepackage{nicefrac}       
\usepackage{microtype}      
\usepackage{bbm}
\usepackage{marvosym}
\usepackage{enumitem}

\algrenewcommand\alglinenumber[1]{\tiny #1:}
\algrenewcommand\alglinenumber[1]{\tiny #1:}
\newif\ifextended

\extendedtrue
\ifextended
\newcommand{\extref}[1]{#1}
\else
\newcommand{\extref}[1]{the extended version of the paper~\citep{wu2023lemur}}
\fi

\ifextended
\newcommand{\ext}[1]{#1}
\else
\newcommand{\ext}[1]{}
\fi
\title{\sys: Think and Reason to Solve Text-to-SQL}

\author{%
  Nina Narodytska\\
  VMware Research\\
  \texttt{n.narodytska@gmail.com} \\
  \And
  Shay Vargaftik \\
  VMware Research\\
  \texttt{shay.vargaftik@broadcom.com} \\
}

\begin{document}

\maketitle

\begin{abstract}
Large Language Models (\(\llm\)s) have made significant progress in assisting users to query databases in natural language. While \(\llm\)-based techniques provide state-of-the-art results on many standard benchmarks, their performance significantly drops when applied to large enterprise databases. 
The reason is that these databases have a large number of tables with complex relationships that are challenging for \(\llm\)s to reason about. We analyze  challenges that \(\llm\)s face in these settings and propose a new solution that combines the power of \(\llm\)s in understanding  questions with automated reasoning techniques to handle complex database constraints. Based on these ideas, we have developed a new framework that outperforms state-of-the-art techniques in zero-shot \ttst on complex benchmarks.
\end{abstract}

\section{Introduction}
Large Language Models (\(\llm\)s) have significantly enhanced AI agents' capacity to assist humans in a variety of important tasks, including co-pilot programming~\citep{copilot,githubcopilot}, program verification~\citep{wu2024lemur,chakraborty2023ranking}, and math problem solving~\citep{zhou2024solving}. One of the fastest-growing areas in this space is the development of \(\llm\)-based assistants for querying SQL databases. In this task, a user poses a question to a database in natural language. The agent's goal is to generate an SQL query that, when executed against the database, answers the user's question. Such assistance enables 
users with different levels of expertise to effectively analyze their data.

Recently, \(\llm\)-based solutions have made significant progress in addressing
the text-to-SQL problem~\citep{GaoWLSQDZ24,li2024codes}. While GPT-based methods have quickly reached near-human performance on academic benchmarks, like Spider~\citep{yu-etal-2018-spider}, they struggle to provide high-quality user assistance on large industrial databases~\citep{sequeda2023benchmark,li2024can}.
One of the core challenges is that industrial databases model many objects with complex relationships between them. To transform a natural language question into an SQL query, the \llm must effectively reason about these intricate relationships, which is highly non-trivial for \llm models.
 Interestingly, we found that \gptfourT can even indicate
in some cases that it needs help with logical reasoning on complex databases. Here is a common \gptfourT output message on a question that requires multiple joins from ACME insurance database~\citep{sequeda2023benchmark}: \emph{`This join may need adjustment based on the actual logic of relating claims to policy coverage details.'}. While we do provide the database schema as part of the input, it is still challenging for LLMs to formally reason about database logic.

In this work, we propose a new \ttst framework, \sys, designed for large databases with complex relationships between objects. Our main underlying idea is to combine the ability of \llm models to effectively relate user questions to database objects with the power of automated reasoning to analyze relationships between these objects.
The \sys workflow consists of three high-level steps. First, upon receiving a user's question, we identify the relevant objects and their attributes in the target database. In the second step, we employ an automated reasoner to build a view that joins the relevant tables based on relational constraints defined by the database schema. This view contains all the necessary information to answer the user's questions.
In the third step, we construct a query targeting this view to produce an answer for the user.
Our contributions are summarized as follows:

\begin{itemize}
\itemsep0em 
    \item We propose a \ttst framework \sys capable of querying large industrial databases. To the best of our knowledge, \(\sys\) is the first framework designed to support logical reasoning 
    in the context of the \ttst problem. 
    \item \(\sys\) offers several advantages:
    \begin{itemize}
    \itemsep0em 
        \item alleviates the need for complex reasoning from a \(\llm\), allowing it to focus on tasks where it currently excels,
        \item supports modeling and reasoning about complex, commonly used design patterns to model relationships, like many-to-many, \stars, and \snow,
        \item its modular workflow allows for effective debugging of failures,  
        \item  performs zero-shot generation and does not require  fine-tuning of {\llm}s.
    \end{itemize}
    \item Our experimental results demonstrate significant performance improvements on several standard benchmarks as well as introduced large benchmarks. 
    We also demonstrate the debugging capabilities of \sys.
\end{itemize}

\section{Motivation}
\label{s:motivation}
\tikzstyle{startstop} = [rectangle, rounded corners, 
minimum width=1cm, 
minimum height=0.5cm,
text centered, 
draw=black, 
fill=white!20]

\tikzstyle{data} = [rectangle, rounded corners, 
minimum width=1.7cm, 
minimum height=1cm,
text width=4em,
text centered, 
draw=white, 
fill=white!30]

\tikzstyle{smalldata} = [rectangle, rounded corners, 
minimum width=0.1cm, 
minimum height=0.2cm,
text width=2em,
text centered, 
draw=white, 
fill=white!30]
\tikzstyle{io} = [trapezium, 
trapezium stretches=true, 
trapezium left angle=70, 
trapezium right angle=110, 
minimum width=3cm, 
minimum height=1cm, text centered, 
draw=black, fill=blue!30]

\tikzstyle{process} = [rectangle, 
minimum width=0.2cm, 
minimum height=1cm, 
text centered, 
text width=4em,
draw=black,dashed, 
fill=gray!10]

\tikzstyle{processsmall} = [rectangle, 
minimum width=0.2cm, 
minimum height=1cm, 
text centered, 
text width=3em,
draw=black,dashed, 
fill=gray!10]

\tikzstyle{processbiga} = [rectangle,
 rounded corners, 
minimum width=3.2cm, 
minimum height=1.8cm, 
text centered, 
text width=5em,
draw=black, 
fill=pink!10]

\tikzstyle{processbigb} = [rectangle,
 rounded corners, 
minimum width=4.6cm, 
minimum height=3.4cm, 
text centered, 
text width=5em,
draw=black, 
fill=red!20]
\tikzstyle{processbigc} = [rectangle,
 rounded corners, 
minimum width=9.3cm, 
minimum height=3.4cm, 
text centered, 
text width=5em,
draw=black, 
fill=orange!10]
\tikzstyle{decision} = [diamond, 
minimum width=3cm, 
minimum height=1cm, 
text centered, 
draw=black, 
fill=pink!30]
\tikzstyle{arrow} = [thick,->,>=stealth]
\tikzstyle{line}     = [draw=black,thick]



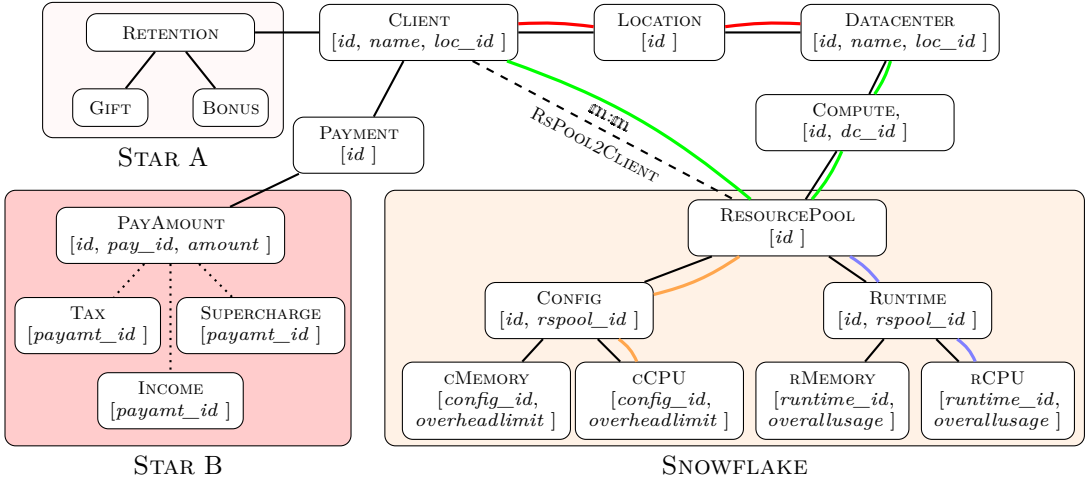
\begin{figure}
\centering

\begin{tikzpicture}[auto, node distance=2cm, every node/.style={sloped}]
\node (retention) [startstop,font=\scriptsize, text width=2cm ]{\rst };
\node (client) [startstop, font=\scriptsize, right of=retention, xshift= 1.3cm, text width=2.4cm]{\client \ \ \ \ \ \ \ \ \ \ \ \ \ \ \ [\id, \name, \locid]};
\node (loc) [startstop, font=\scriptsize, right of=client, xshift= 1.2cm, text width=1.5cm]{\loc \ \ \ \ \  [\id]};
\node (dc) [startstop, font=\scriptsize, right of=loc, xshift= 1.2cm, text width=2.4cm]{\dc \ \ \ \ \ \ \ \ [\id, \name, \locid]};
\node (gift) [startstop, font=\scriptsize, below of=retention, xshift= -0.8cm,yshift= 1cm]{\gift };
\node (bonus) [startstop, font=\scriptsize, below of=retention, xshift= 0.8cm,yshift= 1cm]{\bonus};
\node (pam) [startstop, font=\scriptsize, below of=client, xshift= -0.8cm,yshift= 0.5cm, text width=1.5cm]{\pam \ \ \ \ \  [\id]};
\node (pama) [startstop, font=\scriptsize, below of=retention, xshift= 0cm,yshift= -0.7cm, text width=2.8cm]{\pama\ \ \ \ \ \ \ \ \ \ \  \ \ \ \ \ [\id, \pamid, \amount]};
\node (tax) [startstop, font=\scriptsize, below of=pama, xshift= -1.1cm,yshift= 0.8cm, text width=1.7cm]{\tax [\pamaid]};
\node (charge) [startstop, font=\scriptsize, below of=pama, xshift= 1.2cm,yshift= 0.8cm, text width=2cm]{\charge  [\pamaid]};
\node (income) [startstop, font=\scriptsize, below of=pama, xshift= -0cm,yshift= -0.2cm, text width=1.7cm]{\income  [\pamaid]};
\node (compute) [startstop, font=\scriptsize, below of=dc, xshift= -0.6cm, yshift= 0.8cm, text width=2.4cm]{\compres,\ \ \ \ \ \ \ [\id, \dcid]};
\node (pool) [startstop, font=\scriptsize, below of=compute, xshift= -0.9cm, yshift= 0.6cm, text width=2.4cm]{\respool \ \ \ \ \  [\id]};
\node (config) [startstop, font=\scriptsize, below of=pool, xshift= -2.9cm, yshift= 0.9cm, text width=2cm]{\config \ \ \ \ \ \ \ \  \ \ [\id, \respoolid]};
\node (runtime) [startstop, font=\scriptsize, below of=pool, xshift= 1.6cm, yshift= 0.9cm, text width=2cm]{\runtime \ \ \ \  \ \ \ [\id, \respoolid]};
\node (cmem) [startstop, font=\scriptsize, below of=config, xshift= -1.1cm, yshift= 0.8cm, text width=2cm]{\cmem [\configid, \overheadlimit]};
\node (ccpu) [startstop, font=\scriptsize, below of=config, xshift= 1.2cm, yshift= 0.8cm, text width=2cm]{\ccpu \ \ \ \ \ \ \  \ [\configid, \overheadlimit]};
\node (rmem) [startstop, font=\scriptsize, below of=runtime, xshift= -1cm, yshift= 0.8cm, text width=1.8cm]{\rmem [\runtimeid, \overallusage]};
\node (rcpu) [startstop, font=\scriptsize, below of=runtime, xshift= 1.2cm, yshift= 0.8cm, text width=1.8cm]{\rcpu [\runtimeid, \overallusage]};
\begin{scope}[on background layer]
\node (stara) [processbiga, font=\scriptsize, below of=retention, xshift= -0.1cm, yshift= 1.5cm, label=below:\stars A]{};
\end{scope}
\begin{scope}[on background layer]
\node (starb) [processbigb, font=\scriptsize, below of=pama, xshift= 0.1cm, yshift= 0.9cm, label=below:\stars B]{};
\end{scope}

\begin{scope}[on background layer]
\node (starc) [processbigc, font=\scriptsize, below of=pool, xshift= -0.7cm, yshift= 0.8cm, label=below:\snow]{};
\end{scope}

\draw [line] (retention) --  (client);
\draw [line] (client) --  (loc);
\draw [line] (loc) --  (dc);
\path (client) edge [bend left=5, red, very thick] (loc);
\path (loc) edge [bend left=5, red, very thick] (dc);
\path (pool) edge [bend left=10, orange!70,very thick] (config);
\path (config) edge [bend left=15, orange!70,very thick] (ccpu);
\path (pool) edge [bend left=10, blue!50, very thick] (runtime);
\path (runtime) edge [bend left=15, blue!50,very  thick] (rcpu);

\draw [line] (retention) --  (gift);
\draw [line] (retention) --  (bonus);
\draw [line] (client) --  (pam);
\draw [line] (pam) --  (pama);
\draw[thick,black,dotted](pama) --  (tax);
\draw[thick,black,dotted] (pama) --  (charge);
\draw[thick,black,dotted] (pama) --  (income);
\draw [line] (dc) --  (compute);
\path (dc) edge [bend left=8,green, very thick] (compute);
\path (compute) edge [bend left=8,green,very thick] (pool);
\path (client) edge [bend left=10,green, very thick] (pool);

\draw [line] (compute) --  (pool);
\draw [line] (pool) --  (config);
\draw [line] (pool) --  (runtime);
\draw [line] (config) --  (cmem);
\draw [line] (config) --  (ccpu);
\draw [line] (runtime) --  (rmem);
\draw [line] (runtime) --  (rcpu);
\draw[thick,black,dashed]  (client) --  node [above] {\scriptsize{$\manymany$}} node [below] {\scriptsize{\restoclient}} (pool);

\end{tikzpicture}
\caption{Objects and their relations in the database \cdd.} \label{fig:schema}
\end{figure}
To provide high-quality user assistance in \ttst tasks, we face two types of challenges. The first type of challenge comes from the formulation of the user's question. A question can be poorly specified, ambiguous, or require additional knowledge that is not present in the question. For example, the user might ask to list clients eligible for a loan; however, the eligibility criteria are not present in the question~\citep{li2024can,birdleader}. The second class is related to the complexity of the queried database that can have a large number of tables with complex relations between them~\citep{sequeda2023benchmark, li2024can}.  In this work, we focus on the second class.
One approach to deal with complex relationships is to introduce an intermediate layer,
like a knowledge graph or ontology structure, that contains rich information about the underlying database. Then, LLMs generate queries to this knowledge graph using specialized languages, e.g., SPARQL,~\citep{sequeda2023benchmark}. In turn, these queries can be automatically translated to SQL. While this approach does show promise, it does not alleviate the core issue: an LLM is still expected to reason about complex relations between objects in this intermediate representation. Moreover, such a rich intermediate layer, like an ontology, might not be easy to obtain for a database. Other standard techniques, like additional training, multi-shot or fine-tuning, also rely on {\llm}s to perform constrained reasoning steps~\citep{gao2023texttosql,pourreza2024dtssql,GaoWLSQDZ24}. To the best of our knowledge, dealing with complex relationships in \ttst  remains an open problem.
In order to isolate the underlying challenges in this problem, we created an example database that covers standard relationship  patterns adopted in industry and academia.
 We  identified a set of simple and clearly formulated questions and demonstrated that even on this simplified schema and clear questions, state-of-the-art {\llm}s struggle to assist the user.
\vspace*{-5pt}
\subsection{Database description}
\label{ss:db_description}

We describe a minimal example database schema that contains basic relations, like \oneone and \onemany, and more advanced relationship patterns, like \manymany and \stars,
and analyze the performance of {\llm}s on this  schema (See Appendix~\ref{a:s:backgroud} for relational database definitions). 
%
Suppose a business sells cloud compute resources to customers and uses a database, \cdd, to manage its Day-to-Day Operations. Figure~\ref{fig:schema} shows objects' corresponding tables, their relationships, and a subset of attributes. In particular, each table has a primary key, e.g., $\loc.\id$, and might have foreign keys to refer to another table, e.g., \client refers to \loc using $\client.\locid$. All attributes relevant to our examples are shown in Figure~\ref{fig:schema} with self-explanatory names.
\cdd manages payments (\pam) and marketing retention strategies~(\rst) for clients~(\client) and resources~(\respool) in datacenters~(\dc). This example is in part inspired by the VMware vSphere data model (discussed in Section~\ref{s:exp}).  The full data model contains hundreds of types of resources that form deep tree-like  structures~\citep{vsphereobjects}.
Next, we consider how relationships between objects are modeled in \cdd. Figure~\ref{fig:schema} already defines basic relationships, including \oneone (dotted edges) and \onemany (solid edges).

\noindent\textbf{Many-to-many (\manymany).} \(\client\) and \(\respool\) are related via a \manymany relationship (the dashed edge) meaning that a client might use multiple resource pools and one resource pool can serve multiple clients. The table \restoclient  models this relation.

\noindent\textbf{Star.} A \stars pattern is a type of database schema composed of a single, central fact table surrounded by dimension tables. There are two groups of objects connected in a \stars patterns in our example. \stars~A keeps track of retention marketing strategies for each client that can be either \(\gift\) or/and \(\bonus\). \stars~B records clients' payments (\pam). Payments' amounts are stored in the \pama table. Each amount can be exactly one of three types: \(\tax\), \(\charge\), and \(\income\).

\noindent\textbf{Snowflake.} A \snow schema consists of one fact table connected to many dimension tables, which can be connected to other dimension tables through a many-to-one relationship.
In \cdd, database resource pools are modeled using the snowflake pattern. Each resource pool has configurations (\config) and snapshots of the current usage (\runtime). \config and \runtime have two children nodes each to define CPU and memory properties.

\noindent\textbf{Lookup.} A \lookup table is a table that contains descriptions and code values used by multiple tables, e.g., zip codes, country names. etc. In \cdd, \loc is a lookup table that stores geo-location related data for quick access.

\begin{table}
    \centering
    \scriptsize
    \begin{tabular}{@{}l@{}}
       
      \hline
        \emph{\qone: List customers who use datacenters with names} \\
        \emph{starting with `dev'. Output clients and datacenters}\\
                {\emph{names.}}\\ 
      \hline
\begin{lstlisting}[
          label=lst:q1,
           language=SQL,
           showspaces=false,
            keywordstyle=\scriptsize\color{blue}\ttfamily,
           basicstyle=\scriptsize\ttfamily,
           commentstyle=\color{gray},
        mathescape=true
        ]
/*GPT4 generated SQL*/: 
select $\client.\name$, $\dc.\name$ 
from $\client$ 
join $\textcolor{red}{\loc}$ on $\textcolor{red}{\loc.\id}$ = $\client.\locid$
join $\dc$ on $\textcolor{red}{\loc.\id}$ = $\dc.\locid$
where $\dc.\name$ like 'dev%'
\end{lstlisting} \\
  \hline
\begin{lstlisting}[
          label=lst:a1,
           language=SQL,
            keywordstyle=\scriptsize\color{blue}\ttfamily,
           showspaces=false,
           basicstyle=\scriptsize\ttfamily,
           commentstyle=\color{gray},
                      mathescape=true
        ]
/*Correct SQL*/ 
select $\client.\name$, $\dc.\name$           
from $\dc$
join $\compres$ on $\dc.\id$ = $\compres.\dcid$
join $\respool$     on 
$\quad~$ $\compres.\id$ = $\respool.\compresid$
join $\restoclient$ on
$\quad~$ $\respool.\id$ = $\restoclient.\respoolid$
join $\client$ on $\client.\id$ = $\restoclient.\clientid$
where $\dc.\name$ like 'dev%'
\end{lstlisting}\\
      \hline
      \end{tabular}
      \quad
       \begin{tabular}{@{}l@{}}
            \hline
\emph{ \qtwo: List resource pools names with CPU overhead} \\
{limit greater than runtime overall usage by 100.}\\
      \hline


\begin{lstlisting}[
          label=lst:q2,
           language=SQL,
           showspaces=false,
            keywordstyle=\scriptsize\color{blue}\ttfamily,
           basicstyle=\scriptsize\ttfamily,
           commentstyle=\color{gray},
             mathescape=true
        ]
/*GPT4 generated SQL*/: 
select $\respool.\name$ 
from $\respool$  
join $\rcpu$  on 
$\quad~$ $\respool.\textcolor{red}{\runtimeid}$ =  $\rcpu.\runtimeid$ 
join $\ccpu$  on 
$\quad~$ $\respool.\textcolor{red}{\configid}$ =  $\ccpu.\configid$ 
where $\ccpu.\overheadlimit$ >  $\rcpu.\overallusage$ + 100
\end{lstlisting}
 \\
\hline
\begin{lstlisting}[
          label=lst:a2,
           language=SQL,
            keywordstyle=\scriptsize\color{blue}\ttfamily,
           showspaces=false,
           basicstyle=\scriptsize\ttfamily,
           commentstyle=\color{gray},
           mathescape=true
        ]
/*Correct SQL*/: 
select distinct $\respool.\name$  
from  $\respool$
left join $\config$ on 
$\quad~$ $\respool.\id$ = $\config.\respoolid$
left join $\ccpu$  on $\config.\id$ = $\ccpu.\configid$
left join $\runtime$ on 
$\quad~$ $\respool.\id$ =  $\runtime.\respoolid$
left join $\rcpu$  on  $\runtime.\id$ = $\rcpu.\runtimeid$
where $\ccpu.\overheadlimit$ > $\rcpu.\overallusage + 100$
\end{lstlisting}
\\
      \hline
    \end{tabular}
    \caption{User's questions \qone~and \qtwo. Incorrect parts of the \gpt answer are shown in red.}
    \vspace{-15pt}
    \label{tab:q1q2}
\end{table}
\subsection{User questions}
\label{ss:db_questions}

We consider three simple questions to \cdd that are well formulated: outputs are explicitly specified, so no additional information 
is needed to answer them. We use \gptfourT(`gpt-4-0125-preview'), and \promptB~\citep{sequeda2023benchmark} for these questions. For each question, we present a ground truth answer and a \gpt answer. 
Table~\ref{tab:q1q2} presents both questions (\qthree~ is presented in Appendix~\ref{a:ss:userquestion}). 

Question {\qone} is \emph{`List customers who use datacenters with names starting with `dev'. Output clients and datacenters names'}. 
The user asks for information that relates clients and datacenters. Consider \gpt's answer. 
\(\gpt\) misses the core logic of the database: \emph{clients} and datacenter \emph{resources} are related via a \manymany relation (modeled with  \restoclient). \gpt outputs clients and datacenters that share the same location, which is incorrect.

Question {\qtwo}  is \emph{`List resource pool names with CPU overhead limit greater than runtime overall usage by 100'}. Here the user asks about resource pool properties. However, the \(\gpt\) answer ignores the database's primary/foreign relations. It performs an \ijn between \(\respool\), \(\ccpu\), and \(\rcpu\) tables, using non-existent attributes $\respool.\configid$ and $\respool.\runtimeid$, which is clearly incorrect.

In summary, these examples demonstrated  that 
\llm{}s struggle to handle complex relationships between objects.

\section{Framework design}
In this section, we present our framework \(\sys\).
Figure~\ref{fig:diagram} 
illustrates the workflow diagram, and Algorithm~\ref{algo:lucy} shows the main steps of the workflow. There are two inputs to the framework. The first input is a user question \Question. The second input is \(\dbm\), which is a description of the database schema that we discuss in the next section (Section~\ref{ss:dbm}).
%
The workflow consists of three sequential subtasks: \(\retrieve\), \(\solve\), and \(\tvs\). \(\retrieve\) identifies the relevant tables and their attributes related to the user question (Section~\ref{ss:retrieve}). \(\solve\) finds a combined view of relevant tables taking into account database constraints (Section~\ref{ss:solve}). The third phase, \(\tvs\), takes \(\View\) and the user question \Question and produces an SQL query \(\Query\) for \(\View\) (Section~\ref{ss:tvs}). To simplify notations, we assume that \dbm is a global variable in Algorithm~\ref{algo:lucy}.
 
\tikzstyle{startstop} = [rectangle, rounded corners, 
minimum width=1cm, 
minimum height=1cm,
text centered, 
draw=black, 
fill=purple!20]

\tikzstyle{data} = [rectangle, rounded corners, 
minimum width=0.8cm, 
minimum height=1cm,
text width=2.8em,
text centered, 
draw=black, 
fill=white!30]

\tikzstyle{dataEx} = [rectangle, rounded corners, 
text centered, 
draw=black, 
fill=white!10]

\tikzstyle{io} = [trapezium, 
trapezium stretches=true, 
trapezium left angle=70, 
trapezium right angle=110, 
minimum width=3cm, 
minimum height=1cm, 
text centered, 
draw=black, fill=blue!30]

\tikzstyle{process} = [rectangle, 
minimum width=0.2cm, 
minimum height=1cm, 
text centered, 
text width=4em,
draw=black,dashed, 
fill=green!10]

\tikzstyle{processsmall} = [rectangle, 
minimum width=0.2cm, 
minimum height=1cm, 
text centered, 
text width=3em,
draw=black,dashed, 
fill=green!10]

\tikzstyle{processbig} = [rectangle,
 rounded corners, 
minimum width=2.2cm, 
minimum height=1cm, 
text centered, 
draw=black, 
fill=green!10]

\tikzstyle{decision} = [diamond, 
minimum width=3cm, 
minimum height=1cm, 
text centered, 
draw=black, 
fill=green!30]
\tikzstyle{arrow} = [thick,->,>=stealth]
\tikzstyle{line}     = [draw, -latex']



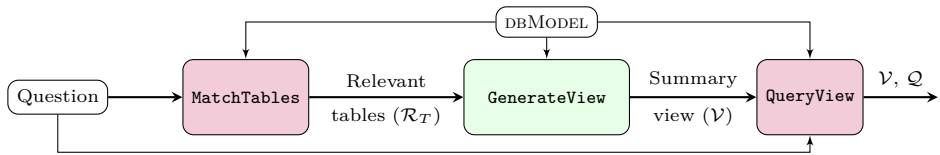
\begin{figure}
\centering

\begin{tikzpicture}[node distance=2cm]
\node (input1) [dataEx, font=\scriptsize]{Question};

\node (retrieve) [startstop,  font=\scriptsize, right of=input1, xshift= 0.5cm] {$\retrieve$};

\node (solveoverall) [processbig,  font=\scriptsize,  right of=retrieve, xshift=2cm ] {\solve};
\node (sql) [startstop, font=\scriptsize,  right of=solveoverall, xshift=1.5cm] {$\tvs$};

\node (input0) [dataEx,  font=\scriptsize,  above of=solveoverall, xshift=0cm,yshift= -1cm]{\dbm};

\draw [arrow] (input1)to[right] node[anchor=south] {} (retrieve);

\draw [arrow] (retrieve) --   node [above] {\scriptsize{Relevant}} node [below] {\scriptsize{tables ($\relta)$}}  (solveoverall) ;
\draw [arrow] (solveoverall) --  node [above] {\scriptsize{Summary}}   node [below] {\scriptsize{view ($\View$)}}  (sql);
\draw [arrow] (sql) to[right] node[above] {\scriptsize{$\View$}, \scriptsize{\Query}}   ++ (1.7,0);
\path[line] (input1.south) -- +(0,-0.5) -|  (sql.south) node[pos=0.25,below]{} ;  
\path[line] (input0.east) -- +(0,0) -|  (sql.north) node[pos=0.25,above]{} ;  

\path[line] (input0.west) -- +(0,0) -|  (retrieve.north) node[pos=0.25,above]{} ;  
\path[line] (input0.south) -- +(0,0) -|  (solveoverall.north) node[pos=0.25,above]{} ;  

\end{tikzpicture}
\caption{\sys's high-level workflow. Red colored boxes indicate phases performed by {\llm}s, and
a green colored box is a phase performed by an automated reasoner.} \label{fig:diagram}
\end{figure}


\subsection{Database model (\dbm)}
\label{ss:dbm}
We start with \dbm, or \dbmshort for short. \dbmshort is a data structure that contains aggregated information about the database, maintained as a JSON structure.  \(\dbmshort\)  should be constructed once for a database as the structure of the database
is relatively stable. \(\dbmshort\) can always be extended if the database requires modifications.
%
Here are the two main blocks of \dbmshort:

\noindent\textit{Database schema.}  
The schema is written using the SQL Data Definition Language
(CREATE TABLE statements). It includes table names, names and types of columns in each table, and database constraints such as primary and foreign keys. 
It can also contain optional user comments associated with each table and column.
We refer to tables and constraints as  \dbmshort.\tables and  \dbmshort.\cons, respectively. 
We extract this information in the form of JSON.
Appendix~\ref{a:sss:dbm_tables}--\ref{a:sss:dbm_cons} shows examples of these structures.

\noindent\textit{Patterns summary.} The user can optionally  list higher-level design patterns that are not captured by the schema explicitly. This information can help to improve the accuracy of the algorithm.
 We support \manymany, \stars, \snow, and \lookup patterns, but  the model is extendable to support other patterns. The user identifies these patterns manually, based on the logic of the target domain. In the future, we envision that the process can be partially automated.
 Appendix~\ref{a:sss:dbm_patterns} shows the JSON format used to specify pattern structures.

\noindent\textbf{Formal notations.}
We introduce formal notations.
\(\dbmshort.\tables\) contains a list of tables $t_i$, $i \in [1,m]$ where $m$ is the number of tables.
\(\dbmshort.\cons\) contains a set of pairs \((t_i, t_j)\) such that \(t_i\) and \(t_j\) are related via \(\oneone\), \(\onemany\) or \(\manyone\) relation. We denote \(\dbmshort.\manymany\) as a set of triplets  \((t_i, t_j, t_k)\), where a join table \(t_k\) models a \(\manymany\) relation between tables \(t_i\) and \(t_j\). Note that \((t_i, t_k)\) and \((t_j, t_k)\) must be in \(\dbmshort.\cons\). Additionally, we denote \(\dbmshort.\lookup\) as the set of lookup tables. For example, in the \(\cdd\) database, \(\dbmshort.\manymany = \{(\client, \respool, \restoclient)\}\) and \(\dbmshort.\lookup = \{\loc\}\).
For a tree-like pattern,  like \(\stars\) or \(\snow\), we distinguish between root table and inner tables using two predicates, e.g., \texttt{star\_root}(\tbl) returns \true if \(\tbl\) is the root table of a \(\stars\) and \texttt{star\_inner}(\tbl) returns \true if \(\tbl\) is an inner table (not root) of a \(\stars\).
\subsection{The \retrieve phase}
\label{ss:retrieve}

The first phase, \(\retrieve\), needs to find relevant tables and their attributes to the user question. One approach to achieve that can be to provide the schema and a question to an \llm and ask for 
this information. 
However, one of the distinguishing features of real-world databases is their large number of tables and attributes. Hence, feeding all of them along with their descriptions to the prompt might not be feasible for many \(\llm\) models. Therefore, we build an iterative procedure that takes advantage of database tree-like patterns. In general, this procedure can be customized to best support the structure of a database.

\begin{algorithm}
  \caption{\sys}\label{algo:lucy}
  \scriptsize
  \begin{algorithmic}[1]
  \Require User question  $\Question$, database model $\dbm$
\Ensure Summary view \View, SQL query $\Query$
  \State\textbf{Phase 1: {\retrieve}  \textcolor{gray}{//\llm-based phase} }
  \State \textcolor{gray}{// get core tables (these are tables that are not inner tables in \stars or \snow)}\label{algo:one:start}
    \State core\_tables = $\{\tbl|\tbl \in \dbmshort.tables \wedge \tbl \notin (\text{snowflake\_inner}(\tbl)\vee \text{star\_inner}(\tbl))\}$ \label{algo:one:core}
  \State  \textcolor{gray}{// identify relevant core tables to the user query}
    \State \_, $T$ = \promptA(\Question, core\_tables, \{\})
\label{algo:one:promptA}
\State $\relta = \{\}$
    \For{$\tbl \in T$} \label{algo:one:explore}
        \If {$\tbl \in \text{snowflake\_root}(\tbl) \vee  \tbl \in \text{star\_root}(\tbl)$}
            \State \textcolor{gray}{// a breadth-first deepening to identify relevant tables and attributes inside a pattern rooted at $t$}
            \State ${\relta} = \relta \cup \text{\ipalgo}(\Question, \tbl)$ \label{algo:one:dive}
        \Else
        \State ${\relta'}, \text{\_}= \promptA(\Question, \{\},\tbl.\attributes)$, ${\relta}  = \relta \cup {\relta'}$  \textcolor{gray}{// identify  $t$'s relevant attributes}
        \label{algo:one:promptA:att}
        \EndIf
    \EndFor\label{algo:one:end}
  \State \textbf{Phase 2: {\solve} \textcolor{gray}{// constraint reasoner-based  phase}}
\State  \textcolor{gray}{// formulate a constraint satisfaction problem}
\State $S = \text{formulate\_csp(\relta)}$ \label{algo:two:graph}
\State  \textcolor{gray}{// solve $S$ to find a path in $G$ that satisfies constraints~\ref{e:relcons}--\ref{e:cost}}
\State $\p = \text{solve\_csp}(S)$ \label{algo:two:path} 
\State  \textcolor{gray}{// build a view  \View base on \p by joining tables along the path $\p$.}
\State $\View = \text{build\_view}(\p)$\label{algo:two:view} 
  \State  \textbf{Phase 3: {\tvs}  \textcolor{gray}{//\llm-based  phase}}
    \State  $\Query$= \promptC(\Question, \View) \label{algo:three:prompt}
    \State \Return $\View, \Query$

  \end{algorithmic}
\end{algorithm}
Algorithm~\ref{algo:lucy} shows \retrieve in lines~\ref{algo:one:start}--\ref{algo:one:end}. First, the algorithm focuses on tables that are not inner tables of any patterns. We refer to such tables as core tables (core\_tables in line~\ref{algo:one:core}). 
For example, Figure~\ref{fig:graph} shows core tables for \cdd. Next, we ask {\llm} to find relevant tables among these core tables using \promptA in line~\ref{algo:one:promptA}.
(Appendix~\ref{a:ss:promptA} shows a \promptA with a few examples.)
As a result, we obtain a set of relevant core tables. We explore them one by one in the loop in line~\ref{algo:one:explore}.
If it is a root table of a pattern, we perform a search inside the corresponding pattern to find more relevant tables using a breadth-first deepening procedure, \ipalgo, in line~\ref{algo:one:dive} (Algorithm~\ref{alg:diving} shows \ipalgo's pseudocode in Appendix~\ref{a:ss:retrieve}). Otherwise, we use \promptA to obtain relevant attributes in line~\ref{algo:one:promptA:att}. 
\begin{example}\label{exm:onetwo:match}
Consider questions 
\(\qone\) and \(\qtwo\)  from Table~\ref{tab:q1q2}. Figure~\ref{fig:graph} shows \cdd's core tables. 
For \qone, a \(\llm\) identifies relevant core tables: T = \{\(\client\), \(\dc\)\} (line~\ref{algo:one:promptA}). 
Since none of these tables is a root of a \(\snow\) or a \(\stars\), we prompt for relevant attributes for each table in line~\ref{algo:one:promptA:att} to get \relta = \{\(\client.\name\), \(\client.\gender\), \(\dc.\name\)\}.
%
%
Now consider \(\qtwo\). 
\llm identifies \(\respool\) as a relevant table in line~\ref{algo:one:promptA}. As \(\respool\) is the root table of  \(\snow\) (see Figure~\ref{fig:schema}), we begin to explore the pattern tree in a breadth-first order using \ipalgo in line~\ref{algo:one:dive}. \(\respool\) has two child nodes, \(\config\) and \(\runtime\), and several attributes. We  query the \(\llm\) and find that both \(\config\) and \(\runtime\) are relevant as well as its attribute \(\respool.\name\). Following the breadth-first search order, we consider \(\config\) with two descendants \(\ccpu\) and \(\cmem\) and discover \(\ccpu\) is relevant
(Example~\ref{a:exm:onetwo:match} in Appendix shows a full version).
\end{example}

\subsection{The \solve phase}
\label{ss:solve}

The \(\retrieve\) phase identifies a set of relevant tables and their attributes. Next, we construct a view table that combines relevant tables and attributes into a single table. 

We build an abstract schema graph \(G\) which provides a graph view of \(\dbmshort\), and define a CSP over this graph. For each table \(\tbl_i\) in \(\dbmshort.\tables\), we introduce a node in \(G\). We use the names \(\tbl_i\) to refer to the corresponding nodes. 
For each pair of tables \(t_i\) and \(t_j\), s.t. \((t_i, t_j) \in \dbmshort.\cons\), we introduce an edge that connects them. We denote \(V\) the set of nodes in \(G\) and \(E\)  its edges. Figure~\ref{fig:graph} illustrates a part of the graph (core tables) for \(\cdd\).
\tikzstyle{startstop} = [rectangle, rounded corners, 
minimum width=1cm, 
minimum height=0.5cm,
text centered, 
draw=black, 
fill=white!20]

\tikzstyle{data} = [rectangle, rounded corners, 
minimum width=1.7cm, 
minimum height=1cm,
text width=4em,
text centered, 
draw=white, 
fill=white!30]

\tikzstyle{smalldata} = [rectangle, rounded corners, 
minimum width=0.1cm, 
minimum height=0.2cm,
text width=2em,
text centered, 
draw=white, 
fill=white!30]
\tikzstyle{io} = [trapezium, 
trapezium stretches=true, 
trapezium left angle=70, 
trapezium right angle=110, 
minimum width=3cm, 
minimum height=1cm, text centered, 
draw=black, fill=blue!30]

\tikzstyle{process} = [rectangle, 
minimum width=0.2cm, 
minimum height=1cm, 
text centered, 
text width=4em,
draw=black,dashed, 
fill=gray!10]

\tikzstyle{processsmall} = [rectangle, 
minimum width=0.2cm, 
minimum height=1cm, 
text centered, 
text width=3em,
draw=black,dashed, 
fill=gray!10]

\tikzstyle{processbiga} = [rectangle,
 rounded corners, 
minimum width=3.2cm, 
minimum height=1.8cm, 
text centered, 
text width=5em,
draw=black, 
fill=pink!10]

\tikzstyle{processbigb} = [rectangle,
 rounded corners, 
minimum width=5.2cm, 
minimum height=1.8cm, 
text centered, 
text width=5em,
draw=black, 
fill=red!20]
\tikzstyle{processbigc} = [rectangle,
 rounded corners, 
minimum width=6.9cm, 
minimum height=2.8cm, 
text centered, 
text width=5em,
draw=black, 
fill=orange!10]
\tikzstyle{decision} = [diamond, 
minimum width=3cm, 
minimum height=1cm, 
text centered, 
draw=black, 
fill=pink!30]
\tikzstyle{arrow} = [thick,->,>=stealth]
\tikzstyle{line}     = [draw=black,thick]



\begin{figure}
\centering

\begin{tikzpicture}[auto, node distance=2cm]
\node (retention) [startstop,font=\small, ]{\rst};
\node (client) [startstop, font=\small, right of=retention, xshift= 0cm]{\client};
\node (loc) [startstop, font=\small, right of=client, xshift= 0.8cm]{\loc};
\node (dc) [startstop, font=\small, right of=loc, xshift= 2.8cm]{\dc};
\node (pam) [startstop, font=\small, below of=client, xshift= -1cm, yshift= 1cm]{\pam};
\node (rel) [startstop, font=\small, below of=client, xshift= 1.8cm, yshift= 1cm]{\restoclient};

\node (compute) [startstop, font=\small, below of=dc, yshift= 1cm]{\compres};
\node (pool) [startstop, font=\small, left of=compute, xshift= -0.5cm]{\respool};


\draw [line] (retention) --  (client);
\draw [line] (client) --  (loc);
\draw [line] (dc) --  (loc);
\draw [line] (client) --  (pam);
\draw [line] (dc) --  (compute);
\draw [line] (compute) --  (pool);
\draw [line] (rel) --  (pool);
\draw [line] (rel) --  (client);
\draw [line] (retention) to[right] node[ anchor=south, below] {...}  ++ (0,-0.4);
\draw [line] (pam) to[right] node[ anchor=south, below] {...}  ++ (0,-0.4);
\draw [line] (pool) to[right] node[ anchor=south, below] {...}  ++ (0,-0.4);

\path (dc) edge [bend left=25,green, very thick] (compute);
\path (compute) edge [bend left=8,green,very thick] (pool);
\path (pool) edge [bend left=8, green, very thick] (rel);
\path (rel) edge [bend left=12,green, very thick] (client);

\path (loc) edge [bend left=12,red, very thick] (client);
\path (dc) edge [bend left=8,red, very thick] (loc);

\end{tikzpicture}
\caption{A part of the abstract schema graph $G$ for \cdd that includes core tables.} \label{fig:graph}
\end{figure}
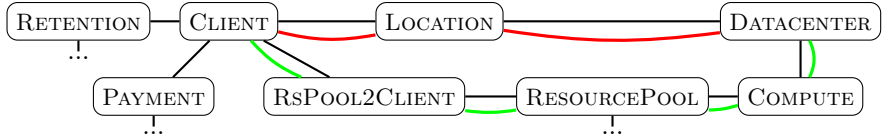

Algorithm~\ref{algo:lucy} shows three main steps of this phase: 
build an abstract graph representation $G$ of the schema (line~\ref{algo:two:graph}); 
formulate and solve CSP to obtain a path \p (line~\ref{algo:two:path}); and 
perform joins along this path to obtain the designed view \View (line~\ref{algo:two:view}). 
Next, we describe these steps.

\noindent\textbf{Problem Formulation.} 
Let \(T = \tables(\relta)\) be a set of relevant tables returned by \(\retrieve\).
We formulate the problem of finding a path \(\p\) in \(G\) that visits a set of nodes \(T\) and satisfies a set of database constraints.
\newcommand{\subscript}[2]{$#1 _ #2$}
\begin{enumerate}[label=(\subscript{C}{{\arabic*}})]
\itemsep-0.1em 
\item \(\p\) must be a valid path in \(G\). This ensures that we
follow primary/foreign keys relationships, i.e., \(\oneone\), \(\onemany\), and 
build a valid view.\label{e:relcons}
\item \(\p\) visits all relevant tables $T$. 
This ensures combining all relevant tables to a view.\label{e:reltables}
\item Consider \((\tbl_i, \tbl_j, \tbl_k) \in \dbmshort.\manymany\). If \(\tbl_i \in \p\) and \(\tbl_j \in \p\) then \(\tbl_k\) must occur in \(\p\) once between \(\tbl_i\) and \(\tbl_j\). These constraints enforce \(\manymany\) relationships. \label{e:m2m}
\item If \(\tbl \in \p\) and \(\tbl \in \dbmshort.\lookupobj\) then \(\tbl\)'s predecessor equals its successor in \(\p\). This ensures that a lookup table serves as a look-up function for each table individually. \label{e:lookup}
\item Cost function: we minimize the number of occurrences of tables outside of \(T\) in \(\p\). A shorter path that focuses on the tables in \(T\) allows us to build more succinct views.\label{e:cost}
\end{enumerate}
\ref{e:relcons}--\ref{e:cost} are common constraints that we encounter in the benchmark sets. In general, the user can specify more constraints to capture the logical relationships of the modeled data.
\noindent\textbf{Constraint satisfaction problem (CSP).} We define a CSP formulation \(S\) of constraints \ref{e:relcons}--\ref{e:cost}. We start with a basic formulation.
Let \(n\) be the maximum length of the path \(\p\). For each node \(t_i\) in \(G\) and step \(r\), where \(r \in [1, n]\), we introduce a Boolean variable \(b_{i}^r\). \(b_{i}^r\) is true iff \(t_i\) is the \(r\)th node in \(\p\). We also introduce a sink-node Boolean variable \(b_{d}^r\) for each layer to model paths that are shorter than \(n\).
\(S\) contains the following logical  constraints:
\begin{align}
\text{\ref{e:cost}}:&& minimize \textstyle\sum\mathop{}_{\mkern-1mu {i, \tbl_i \notin T}}  occ_i && \label{c:cost}\\
    &&\forall i.\tbl_i \in V && \occ_i = b_{i}^1+\ldots + b_{i}^n \label{c:aux:occ}\\
    \text{\ref{e:relcons}}:&& \forall i.\tbl_i \in V, r \in [1,n-1] &&  b_i^r \Rightarrow (\vee_{j.(\tbl_i, \tbl_j) \in E} b_j^{r+1})\vee b_d^{r+1}\label{c:relcons}\\
    \text{\ref{e:reltables}}:&&\forall i.\tbl_i \in T &&  \occ_i \geq  1 \label{c:reltables}
 \end{align}
\begin{align}
    \text{\ref{e:m2m}}:&& \forall k.(\tbl_i, \tbl_j, \tbl_k) \in \dbmshort.\manymany &&  \occ_k = 1 \label{c:m2m1}\\
    \text{\ref{e:m2m}}:&& \forall k.(\tbl_i, \tbl_j, \tbl_k)  \in \dbmshort.\manymany, r \in [2,n-1] &&   b_{k}^r \Rightarrow (b_{i}^{r-1} \wedge b_{j}^{r+1}) \vee  (b_{j}^{r-1}\wedge b_{i}^{r+1})\\ 
    \text{\ref{e:lookup}}:&&  \forall i.\tbl_i \in \dbmshort.\lookupobj, r \in [2,n-1] &&   b_{i}^r \Rightarrow  (b_{j}^{r-1} \Rightarrow b_{j}^{r+1}) \label{c:lookup} \\
    && \forall r \in [1,n] &&  b_{1}^{r} +\ldots + b_{|V|}^{r}= 1\label{c:layer}\\
    && \forall r \in [1,n-1] &&  b_d^r \Rightarrow b_d^{r+1} \label{c:dummy}
\end{align}

Consider the encoding \(S\). Equations~\ref{c:aux:occ} specify integer variables, \(\occ_i\), for \(i \in [1,n]\), that count the occurrences of each table in the path. 
Equations~\ref{c:layer} encode that only one node belongs to a path at each step.
Equations~\ref{c:dummy} encode that if the path visits the sink node, then it must stay there. Other equations encode constraints~\ref{e:relcons}--\ref{e:cost}. By construction, Equations~\ref{c:cost}--\ref{c:dummy} generate a valid path in \(G\) that satisfies the constraints~\ref{e:relcons}--\ref{e:cost}.

\begin{example}
\label{exp:two:view}
For \qone, solving \(S\) gives the green path between \(\dc\) and \(\client\) in Figure~\ref{fig:graph}.  \(S\) rules out the red path as we enforce constraint~\ref{e:lookup} and  optimization~\ref{e:cost}.
\end{example}

\noindent\textbf{Improvements of CSP.}
Our basic model \(S\) can be improved to take advantage of \(\stars\) and \(\snow\) patterns. Namely, we can leverage the decomposition of \(G\)
and find a path \(\p\) among core tables only. Then, for each core table in \(\p\) that is a pattern root, and for each inner relevant table in this pattern, we build a path \(\p'\) along the corresponding branch. For example, Figure~\ref{fig:schema} shows two paths from \(\respool\) to \(\ccpu\) (an orange path) and \(\rcpu\) (a blue path). We use \(\ljn\) to combine tables along each such branch. Finally, we combine \(\p\) and {\p'}s into a single view.

\noindent\textbf{Summary view.}
Given a path $\p$ in a graph, we \jn tables along the path using their primary and foreign key relations. 
We keep the same set of attributes that \retrieve identified.
An example of the \View for \qone~ that corresponds to the green path in Figure~\ref{fig:graph} is shown in the listing in Table~\ref{tab:q4} in Appendix~\ref{a:ss:solve:view}.

\subsection{The \tvs phase.}
\label{ss:tvs}
\(\tvs\) takes the summary view \(\View\) along with the user question, and prompts an \(\llm\) to obtain the final SQL using \promptC (line~\ref{algo:three:prompt} in Algorithm~\ref{algo:lucy}). \promptC is defined in Appendix~\ref{a:ss:promptC}. The listing in Table~\ref{tab:q4} shows an SQL \(\Query\) to answer \qone~(Appendix~\ref{a:ss:solve:view}).

\section{Discussion on strengths and limitations }
\label{ss:discussion}
\noindent\textbf{Strengths.}
\(\sys\) is designed based on the principle of separation of responsibilities between generative tasks and automated reasoning tasks: each step focuses on either an NLP-related subproblem or a constraint reasoning subproblem. This separation allows us to support a number of unique capabilities.
First, \(\sys\) shifts the burden of complex reasoning from \(\llm\)s to constraint solvers. 
%
Second, we support reasoning on complex relationships,
like \manymany, \lookup, \stars or \snow. 
%
Third, our framework is flexible and extensible as it is easy to incorporate domain-specific constraints as soon as they can be expressed by constraint modeling language. This assumes that the user has a data analytics role and understands the logic of the database.  Such formal reasoning capability
is important, as it is hard to control \(\llm\)s via prompts when non-trivial reasoning is required.
Fourth, we can  evaluate each phase and diagnose \(\sys\) failure modes. For example, if \(\retrieve\) misses relevant tables, this indicates that we need to provide more information about the schema to an \(\llm\).
Fifth, based on our  evaluation, \sys can support complex queries that include multiple filtering operators and aggregators, e.g. average or sum. This capability follows from the \tvs phase as the final call to an \(\llm\) is performed on a single view table.

\noindent\textbf{Limitations.}
The first limitation is that we cannot guarantee that the SQL query answers the user's question.
Given the current state of the art, providing such guarantees is beyond the reach of any copilot method that takes natural language descriptions  and outputs structured text, like code or SQL. However, our solution does guarantee that  \View  satisfies  database constraints, which is a step forward in this direction.
Second, we do not support questions that require \union operators in the \solve phase. In fact, there are no benchmarks available that require the \union operator to answer questions. Supporting \union would require an extension of \retrieve and \solve.
Third, we observed experimentally that \sys struggles with certain types of queries that involve a particular interleaving ordering of filtering and aggregate operators or question-specific table dependencies, like a lookup table that has to be used multiple times to answer the user's question. We further discuss such questions in our experiments.
\section{Experimental evaluation}
\label{s:exp}
In our experimental evaluation, we aim to answer the main questions:
\begin{itemize}
\itemsep0em 
    \item Is \(\sys\) competitive with existing \(\llm\)-based approaches? 
    \item Can we  debug \(\sys\) to gain insights about failure modes? 
    \item Can \(\sys\) handle complex questions? 
\end{itemize}

\noindent\textbf{Setup.}
We compare with the following zero-shot baselines: \gptfourT, \nsql, and \chat (\chatfour for short). \gptfourT and \chatfour methods are the best zero-shot techniques according to the BIRD leadership board that are accessible for evaluation~\citep{birdleader}. \nsql is the best open-source large foundation model designed specifically for the SQL generation task \citep{numbersstation2023NSText2SQL}. \chat is closed-source but the authors kindly extended their API that we can run experiments with \gptfourT.  We provide all benchmarks and frameworks' results in the supplementary materials. For \gptfourT and \sys, we use the ‘gpt-4-0125-preview’ API without fine-tuning. We use OR-Tools as a constraint solver~\citep{cpsatlp} (Appendix~\ref{a:exp:setup} provides full details of the experimental setup).
%

\noindent\textbf{Evaluation metrics.} 
We use the standard Execution Accuracy (\ex)~\citep{li2024can}. In addition, we consider a relaxation of this metric. We noticed that frameworks often add additional attributes to the output as the exact format of the output is rarely specified.
Hence, we extend \ex to \(\esx\) metrics that check if the output of a framework contains the ground truth outputs. To better understand performance characteristics 
and 
possible failure modes, we consider the coverage metric that captures whether a framework correctly identified a subset of relevant tables and attributes. Let \(sql_G\) be the ground truth answer and \(sql_F\) be a generated query. Then we assess the percentage of the ground truth content \(slq_F\) captures:
\begin{align}
\cover_\tbl = \frac{|\tables(slq_F) \cap \tables(slq_G)|}{|\tables(slq_G)|} &
~~\cover_\att = \frac{|\attributes(slq_F) \cap \attributes(slq_G)|}{|\attributes(slq_G)|},
\end{align}
where \tables() and \attributes() are functions that return a set of tables and attributes.
\paragraph{ACME insurance.}
\begin{table}[!htb]
                \begin{minipage}{.5\linewidth}
    \centering
    \vspace{-10pt}
\caption{\label{tab:insurance_gt} The \insurance dataset.} 
                \scriptsize                \centering 
                 \begin{tabular}{|l|rrrrrr|}
\hline 
 &\gptfourT&\gptfourTS&\chatfour&\nsql&\sys& \dataw\\ 
 \hline 
\hline 
$\cover_\tbl$ & 0.44 & 0.47 & 0.82 & 0.31 & \textbf{0.95} &-\\ 
$\cover_\att$ & 0.36 & 0.42 & 0.81 & 0.25 & \textbf{0.93} &-\\ 
 \hline 
$\ex$ & 9 & 13 & 16 & 2 & \textbf{30} & 24\\ 
$\esx$ & 9 & 13 & 16 & 3 & \textbf{33} &-\\ 
 \hline 
                \end{tabular}  
  \vspace{-5pt}
\end{minipage}\hfill
\begin{minipage}{.5\linewidth}
    \centering
      \vspace{-10pt}
\caption{\label{tab:vsphere_large_gt} The \vlarge dataset.} 
                \scriptsize                \centering 
                 \begin{tabular}{|l|rrrr|}
\hline 
 &\gptfourT&\gptfourTS&\chatfour&\sys\\ 
 \hline 
\hline 
$\cover_\tbl$ & 0.46 & 0.44 & 0.44 & \textbf{0.98} \\ 
$\cover_\att$ & 0.50 & 0.44 & 0.48 & \textbf{0.98} \\ 
 \hline 
$\ex$ & 6 & 4 & 2 & \textbf{17} \\ 
$\esx$ & 9 & 5 & 2 & \textbf{18} \\ 
 \hline 
                \end{tabular}   
                  \vspace{-5pt}

\end{minipage}
\end{table}
We consider the \insurance dataset that was recently published~\citep{sequeda2023benchmark}. The dataset represents an enterprise relational database schema in the insurance domain. The authors focused on a subset of 13 tables out of 200 tables and proposed a set of 45 challenging questions.
We identified two \stars patterns in this database. The authors showed that their method (\dataw) solved 24 out of 45 problems using intermediate representation of a knowledge graph, while \gptfourT solved only 8 problems. 
However, results are not publicly available, so we cannot perform coverage analysis and compute \esx. 

We reran the experiment on \(\gptfourT\) with the same \promptB (Appendix~\ref{a:ss:promptB}) and obtained similar results to those reported in~\citep{sequeda2023benchmark}. In addition, we extended the schema with descriptions of table attributes from \dbm in the form of comments, which we called \(\gptfourTS\) (See Appendix~\ref{a:ss:insurance} for examples). Table~\ref{tab:insurance_gt} shows our results. First, we observe that there is a strong correlation between coverage and accuracy metrics in the results.
\chatfour and \sys show good coverage, meaning that they can correctly identify most of the required tables and attributes. They also demonstrate better performance compared to other methods. Our framework shows very high coverage and solves about 30 of benchmarks according to the \ex metric, which outperforms \dataw that solves 24 and 
other methods.

\sys still cannot solve 13 benchmarks, which is surprising given  high coverage.
We performed a study to locate where \(\sys\) fails on these benchmarks (See Appendix~\ref{a:sss:insurance} for all questions where \(\sys\) was unsuccessful). In summary, the majority of failures come from under-specified output attributes or nonstandard aggregators, like specialized formulas to compute an average. In four cases, \(\retrieve\) missed a table, and in one case, \(\tvs\) missed the attribute to output. The most interesting mode of failure is when we need to perform multiple lookups on the same table. 
The reason for that is the \retrieve phase identifies only relevant tables but ignores possible relationships between them. Extending \retrieve to retrieve relationships between tables is interesting future work.

\paragraph{BIRD datasets.}
Next, we consider the state-of-the-art dataset BIRD~\citep{li2024can}. From the development set, we chose two datasets with complex relationships between objects: \financial (106 instances) and \formula (174 instances)\footnote{Recently, \cite{wretblad2024understanding} provided a detailed analysis of the BIRD dataset and found a  number of errors of various types. See Appendix~\ref{a:ss:bird} for the discussion.}. The accuracy of \chat on the BIRD development set is $\sim58\%$; however, its accuracy on \financial and \formula are much lower, $\sim45\%$. 
We compare with results from \gptfour and \chatfour available from~\citep{gptbird} and ~\citep{chat2querybench}, respectively. However, we reran these benchmarks with \gptfourT and \gptfourTS as the \gptfour results are nearly one year old. 
%
Table~\ref{tab:financial_gt-corrected} and Table~\ref{tab:formula_1_gt-corrected} show results on \financial and \formula, respectively. 
\(\sys\) and \(\chatfour\) have higher coverage and good accuracy. \(\sys\) shows the best results in most cases. Again, \(\sys\) has very good coverage on \financial but was able to solve only 68 out of 106 queries based on the \(\esx\) metric. We manually performed an questions study on the failed questions. There are two major groups there that are interesting. First, \(\sys\) has difficulty if there are multiple orderings, especially nested or ordering in different directions. Second, sometimes, \retrieve adds an additional table that is not needed to find the answer.  The rest are either ambiguous questions or small mistakes like outputting a wrong attribute, i.e., \id instead of \name. See Appendix~\ref{a:sss:financial} for examples of questions where \(\sys\) was unsuccessful.
\begin{table}[!htb]
                \begin{minipage}{.5\linewidth}
    \centering
  \vspace{-10pt}
 \caption{\label{tab:formula_1_gt-corrected} The \formula dataset.} 
                \scriptsize                \centering 
   \begin{tabular}{|l|@{}r@{}r@{}r@{}r@{}r@{}r|}
\hline 
 &~~\gptfour&~~\gptfourT&~~\gptfourTS&~~\chatfour&~~\nsql&~~\sys\\ 
 \hline 
\hline 
$\cover_\tbl$ & 0.86 & 0.78 & 0.77 & 0.88 & 0.52 & \textbf{0.93} \\ 
$\cover_\att$ & 0.84 & 0.75 & 0.75 & 0.81 & 0.50 & \textbf{0.94} \\ 
 \hline 
$\ex$ & 54 & 67 & 65 & 80 & 9 & \textbf{83} \\ 
$\esx$ & 66 & 80 & 79 & 93 & 10 & \textbf{103} \\ 
 \hline 
                \end{tabular}   
\vspace{-5pt}
\end{minipage}\hfill
\begin{minipage}{.5\linewidth}
    \centering
      \vspace{-10pt}
  \caption{\label{tab:financial_gt-corrected} The \financial dataset.} 
                \scriptsize                \centering 
                 \begin{tabular}{|l|@{}r@{}r@{}r@{}r@{}r@{}r|}
\hline 
 &~~\gptfour&~~\gptfourT&~~\gptfourTS&~~\chatfour&~~\nsql&~~\sys\\ 
 \hline 
\hline 
$\cover_\tbl$ & 0.81 & 0.84 & 0.87 & 0.92 & 0.50 & \textbf{0.97} \\  
$\cover_\att$ & 0.81 & 0.81 & 0.85 & 0.91 & 0.59 & \textbf{0.96} \\ 
 \hline 
$\ex$ & 36 & 47 & 52 & \textbf{59} & 6 & 56 \\ 
$\esx$ & 38 & 55 & 64 & 62 & 6 & \textbf{68} \\ 
 \hline 
                \end{tabular}
\vspace{-5pt}
\end{minipage}
\end{table}


\paragraph{Cloud resources.}
Next, we propose a new benchmark based on the vSphere API data model~\citep{vsphereapi}. We experimented with this publicly available data model of an industrial product, as it is well-documented and easily accessible via a web interface. It describes the state of the system as well as its configuration parameters. States are stored in a database and queried by customers to keep track of performance, maintenance, and data analysis. We extracted the descriptions of main objects in~\cite{vsphereobjects}, including data centers, resource pools, hosts, and virtual machines and their properties, and built a database that captures these relationships using 52 tables. 
Overall, we have two {\stars}s, five {\snow}s and two {\manymany}s patterns. For each table and an attribute, we get descriptions from~\citep{vsphereobjects}. As these can be a lengthy description, we use \gpt to shorten it to 15 words (see \promptD in Appendix~\ref{a:ss:promptD}) . We generated data randomly using sqlfaker~\citep{sqlfaker}. We create 20 challenging questions for this benchmark. 

Table~\ref{tab:vsphere_large_gt} shows our results. \nsql cannot process this benchmark due to a limited context window. We again see that \(\sys\) outperforms other models in both coverage and accuracy. \chatfour failed on 6 questions with an error `Unable to generate SQL for this database due to its extensive tables' and it often does not follow instructions on the output columns. In terms of failure mode, \(\sys\) failed in the third phase as it hallucinated some attribute names when names are long, e.g., `Resourcepool\textcolor{red}{runti}memory' instead of `Resourcepool\textcolor{blue}{runtime}memory'.


\bibliographystyle{abbrvnat}
\bibliography{lit}

\newpage
\appendix

\colorlet{punct}{red!60!black}
\definecolor{background}{HTML}{EEEEEE}
\definecolor{delim}{RGB}{20,105,176}
\colorlet{numb}{magenta!60!black}
\lstdefinelanguage{json}{
    basicstyle=\normalfont\ttfamily,
    numbers=left,
    numberstyle=\scriptsize,
    stepnumber=1,
    numbersep=8pt,
    showstringspaces=false,
    breaklines=true,
    frame=lines,
    backgroundcolor=\color{background},
    literate=
     *{0}{{{\color{numb}0}}}{1}
      {1}{{{\color{numb}1}}}{1}
      {2}{{{\color{numb}2}}}{1}
      {3}{{{\color{numb}3}}}{1}
      {4}{{{\color{numb}4}}}{1}
      {5}{{{\color{numb}5}}}{1}
      {6}{{{\color{numb}6}}}{1}
      {7}{{{\color{numb}7}}}{1}
      {8}{{{\color{numb}8}}}{1}
      {9}{{{\color{numb}9}}}{1}
      {:}{{{\color{punct}{:}}}}{1}
      {,}{{{\color{punct}{,}}}}{1}
      {\{}{{{\color{delim}{\{}}}}{1}
      {\}}{{{\color{delim}{\}}}}}{1}
      {[}{{{\color{delim}{[}}}}{1}
      {]}{{{\color{delim}{]}}}}{1},
}

\section{Background}
\label{a:s:backgroud}
\paragraph{Relational databases.} 
Let $D_1,\ldots,D_n$ be a set of domains.
A relation or table, $\tbl$, is defined over subset of domains:
$\tbl(X_{i_0}, \ldots, X_{i_k}) \subseteq D_{i_0}\times \ldots \times D_{i_k}$, $X_{i_j} \subseteq D_{i_j}, j \in [0,k]$. In addition, $\tbl$ defines a set of attributes (or columns) names $X_{i_0}, \ldots, X_{i_k}$.
Projection is a unary operation on a set of attribute names $Y$, $Y \subseteq X$. The result of such projection is the set of tuples that is obtained when all tuples in \tbl are restricted to attributes $Y$. 
Inner join, or simply $\jn$, is a binary operator between two tables $\tbl_1$ and $\tbl_2$ over their common attributes $Y$ that returns a set of all combinations of tuples in $\tbl_1$ and $\tbl_2$  that are equal on $Y$. Left join,
$\ljn$, is similar to the join but returns all rows of $\tbl_1$ filling unmatched rows of $\tbl_2$ with null values.
A database can support a large set of constraints over tables. The two main constraint types are related to primary and foreign keys.
A primary key is the smallest subset of attributes guaranteed to uniquely differentiate each tuple in a table. A foreign key is a subset of attributes $Y$ in a table $\tbl_1$ that corresponds with (usually) a primary key of another table $\tbl_2$, with the property that the projection of $\tbl_1$ on $Y$ is a subset of the projection of $\tbl_2$ on $Y$~\citep{Learning}. 


\paragraph{Design patterns.}
A database typically represents entities and their interactions in real-world processes, e.g., the financial management of a company. To effectively model these complex entities, several design patterns have been developed~\citep{dimensional,patternsbook}.
A many-to-one pattern (\manyone) specifies a relationship when any number of attributes from one table is associated with unique attributes of the same or another table, typically enforced by foreign key and primary key relationships. A many-to-many relationship (\manymany) occurs when any number of attributes from one table is associated with any number of attributes from the same or another table. It is typically modeled with an auxiliary join table that refers to the primary keys of the tables in the relationship.
The \lookup table is a table that contains descriptions and code values used by multiple tables, e.g., zip codes, country names. etc.
A \stars pattern is a type of relational database schema composed of a single, central fact table surrounded by dimension tables. A \snow schema consists of one fact table connected to many dimension tables, which can be connected to other dimension tables through a many-to-one relationship.

\paragraph{Constraint satisfaction.}
A constraint satisfaction problem (CSP) consists of a set of variables, each with a finite domain of values, and a set of constraints specifying allowed combinations of values for subsets of variables~\citep{csp}. A solution is an assignment of values to the variables satisfying the constraints. In the constraint optimization problem, we are looking for a solution that optimizes a given cost function. Constraint solvers typically explore partial assignments enforcing a local consistency property using either specialized or general-purpose propagation algorithms and employ conflict-driven learning to store information from failures as the search proceeds. We used OR-Tools CP-SAT solver~\citep{cpsatlp} in our experiments.

\section{Related work}
\label{s:relatedwork}
We focus on the zero-shot text-to-SQL problem, which has received significant attention in the last few years. \cite{liu2023comprehensive} performed a comprehensive evaluation of ChatGPT on the Spider dataset and demonstrated that it shows good performance. In~\citep{dong2023c3}, a new framework based on the GPT model was proposed, involving several techniques for promoting and post-processing the output to get more consistent results. \cite{chang2023prompt} proposed several techniques to improve the performance of ChatGPT. \citep{chat2querybenchAPI} represents the most recent zero-shot method. According to the API documentation~\citep{chat2querybenchAPI}, the authors construct a data summary object that contains `AI exploration information of the given database.' This method performs very well on the BIRD dataset. However, it relies on {\llm}s to reason about database relationships. \cite{sequeda2023benchmark} performed an interesting investigation of the performance of {\llm}s on large industrial databases. They identified that \gpt does not perform well when it needs to reason about complex relationships. The authors proposed a two-step approach to tackle this problem. As a knowledge graph is available for these benchmarks, the authors proposed using the knowledge graph as an intermediate representation. Namely, the user's question is answered using the \kg structure with SPARQL, and this answer is automatically translated to SQL using a given mapping from ontology to SQL (R2RML). However, while reasoning on a knowledge graph can be easier for {\llm}s, it is still challenging to take all complex relationships into account.
 
\section{Motivation (additional materials)}
\subsection{User's questions}
\label{a:ss:userquestion}
The third question \qthree, \emph{`What are the total tax payments, which is the sum of Tax and Supercharge?'}, asks about the total amount of taxes paid from all payments (Table~\ref{tab:q3}). There are a few issues with the \gpt answer. First, it outputs all payment amounts that are both tax and supercharge. We reminded that
that each payment amount can be of one type, so the result will be empty. Second, it hallucinates as there are no \amount columns in the \(\tax\) or \(\charge\) tables.
\begin{table}
    \centering
    \begin{tabular}{|l|}
            \hline
\emph{\small \qthree: What are the total tax payment, which is the sum of Tax and Supercharge?}\\
      \hline
\begin{lstlisting}[
          label=lst:q3,
           language=SQL,
            keywordstyle=\small\color{blue}\ttfamily,
           showspaces=false,
           basicstyle=\small\ttfamily,
           commentstyle=\color{gray},
                      mathescape=true
        ]
/*GPT4 generated SQL*/: select sum($\tax.\textcolor{red}{\amount}$ + $\charge.\textcolor{red}{\amount}$)
from $\tax$ 
join $\pama$ on $\textcolor{red}{\pama.\id}$ = $\textcolor{red}{\tax.\pamaid}$
join $\charge$ on $\textcolor{red}{\pama.\id}$ = $\textcolor{red}{\charge.\pamaid}$
\end{lstlisting}\\
\hline
\begin{lstlisting}[
          label=lst:a3,
           language=SQL,
            keywordstyle=\small\color{blue}\ttfamily,
           showspaces=false,
           basicstyle=\small\ttfamily,
           commentstyle=\color{gray},
           mathescape=true
        ]
/*Correct SQL*/: 
select sum(ifnull($\ta.\amount$, 0)) + sum(ifnull($\sca.\amount$,0))
from $\pam$
join $\pama$ as $\ta$ on $\pam.\id$ = $\ta.\pamid$
left join $\tax$ on $\ta.\id$ = $\tax.\pamaid$ 
join $\pama$ as $\sca$ on $\pam.\id$ = $\sca.\pamid$
left join $\charge$ on $\sca.\id$ = $\charge.\pamaid$ 
\end{lstlisting}
\\
      \hline
    \end{tabular}
 \caption{A user's question \qthree. Incorrect parts of the \gpt answer are highlighted in red.}
 \label{tab:q3}
\end{table}

\subsubsection{Definition of \promptB}
\label{a:ss:promptB}

\begin{tcolorbox}[colback=white]
\paragraph{Inputs:}  DB\_SCHEMA, Question
\paragraph{\promptB}
Given the database described by the following DDL: <DB\_SCHEMA>.
Write a SQL query that answers the following question. Do not explain the query. Return just the query, so it can be run verbatim from your response. 
Here’s the question: <Question>.~\citep{sequeda2023benchmark}
\paragraph{Returns}: SQL
\end{tcolorbox}

\section{Framework design (additional materials)}
\subsection{Database model (\dbm)}
\subsubsection{Example of a table from \dbmshort.\tables}
\label{a:sss:dbm_tables}

Here is a JSON structure for the Client table from the \financial dataset~\citep{li2024can}. It contains the table name, primary keys, attributes, their types, and descriptions. This information is available in the dataset. The description of the table is generated by \gptfourT using the prompt \promptD.
\begin{lstlisting}[language=json,firstnumber=1]
"Client": {
    "type": "ManagedObject",
    "primary": [
        "client_id"
    ],
    "path": "<path-to>/Client.json",
    "path_to_types": ""<path-to>/Client_types.json"
}
\end{lstlisting}

Here is the JSON structure for Client.json:
\begin{lstlisting}[language=json,firstnumber=1]
{
     "NameField": "Client",
     "DescriptionField": "Focuses on client information, encompassing unique client identifiers, gender, birth dates, and the location of the branch with which they are associated.",
     "client_id": "the unique number",
     "gender": "  Description: 'F: female; M: male '",
     "birth_date": "birth date",
     "district_id": "location of branch"
}
\end{lstlisting}
Here is the JSON structure for Client\_types.json:
\begin{lstlisting}[language=json,firstnumber=1]
{
     "NameField": {
          "type": "varchar(100)",
          "default": "DEFAULT NULL"
     },
     "DescriptionField": {
          "type": "varchar(5000)",
          "default": "DEFAULT NULL"
     },
     "client_id": {
          "type": "bigint",
          "default": "NOT NULL"
     },
     "gender": {
          "type": "varchar(46)",
          "default": "NOT NULL"     
          },
     "birth_date": {
          "type": "date",
          "default": "NOT NULL"
     },
     "district_id": {
          "type": "bigint",
          "default": "NOT NULL"
     }
}
\end{lstlisting}
\subsubsection{Example of a \manyone relation  from \dbmshort.\cons}
\label{a:sss:dbm_cons}
Here is the JSON structure for the Client and District relation from the \financial dataset~\citep{li2024can}.
\begin{lstlisting}[language=json,firstnumber=1]
  "Client, District": {
          "type": "Relationships",
          "sqlrelation": "M:1",
          "foreign_relation": {
               "FOREIGN": [
                    "district_id"
               ],
               "foreign_relation_ref_table": "District",
               "foreign_relation_ref_table_keys": [
                    "district_id"
               ]
          }
     }
\end{lstlisting}

\subsubsection{Example of a \manymany pattern from \dbmshort.\patterndb}
\label{a:sss:dbm_patterns}
Here is the JSON structure for the Account and District \manymany relation from the \financial dataset~\citep{li2024can}.
\begin{lstlisting}[language=json,firstnumber=1]
{
  "Account, Client": {
          "type": "Relationships",
          "description": "",
          "sqlrelation": "M:M",
          "m2m_relation": {
               "m2m_middle_table": "Disp",
               "m2m_side_tables": [
                    "Client",
                    "Account"
               ],
               "m2m_relation_one": [
                    "Disp",
                    "Client"
               ],
               "m2m_relation_two": [
                    "Disp",
                    "Account"
               ]
          }
     }
}
\end{lstlisting}

Here is the JSON structure for the \snow pattern rooted ta \respool (\vlarge dataset).
\begin{lstlisting}[language=json,firstnumber=1]
{
     "NameField": "ResourcePool",
     "config": {
          "cpualloc",
          "memalloc"
     },
     "runtime": {
          "cpu",
          "memory"
     }
}
\end{lstlisting}

\subsection{\retrieve}
\label{a:ss:retrieve}

\subsubsection{\promptA}
\label{a:ss:promptA}
\promptA requires three inputs: a user question, a set of tables  (can be empty), and a set of attributes for a given table $\tbl$ (can be empty). 
\begin{tcolorbox}[colback=white]
\paragraph{Inputs:}  Question, Tables, Attributes 
\paragraph{\promptA:}
Here is a json schema. Please treat json schema objects as a description of tables in a database <JSON(Tables, Attributes)>.
The user has a query to answer <Question>. What are all relevant json elements to a user query from the list [<list of json elements>]?
Output is a list of elements, [element, element,  element,...]. Do not explain.  
\paragraph{Returns:} We post-process the output to extract a set of tables and their attributes (\relta) and relevant tables $T$ 
\end{tcolorbox}
In the prompt, we  provide description of tables and attributes from \dbmshort.
We show a few examples of <JSON(Tables, Attributes)> and the corresponding <list of json elements>.

\begin{example}
Here is an example of a JSON(Tables, \{\}) used in line~\ref{algo:one:promptA} in Algorithm~\ref{algo:lucy} for  \financial dataset. The goal is to determine relevant core tables.

\begin{lstlisting}[language=json,firstnumber=1]
{
     "Account": "Manages financial accounts, tracking each account's unique identification, the location of the associated bank branch, the frequency of account servicing, and the account's creation date. It categorizes the servicing frequency with options like monthly, weekly, and post-transaction issuances Properties of Account: account_id, district_id, frequency, date. ",
     "Card": "Manages of credit cards, incorporating unique identifiers for each card and the related dispositions. It also categorizes credit cards into various classes, such as junior, standard, and high-level, reflecting their tier and associated benefits. Properties of Card: card_id, disp_id, type, issued. ",
     "Client": "Focuses on client information, encompassing unique client identifiers, gender, birth dates, and the location of the branch with which they are associated. Properties of Client: client_id, gender, birth_date, district_id. ",
     "Disp": "Manage dispositions in financial accounts. It contains a unique identifier for each record, links each disposition to specific clients and accounts, and categorizes the nature of each disposition into types like 'OWNER', 'USER', or 'DISPONENT'. Properties of Disp: disp_id, client_id, account_id, type. ",
     "District": "Provides a detailed overview of district-level data, essential for regional analysis and decision-making. It includes a unique identifier for each district, along with the district's name and its broader region. The table delves into demographic, economic data and  economic indicators, records crime statistics. Properties of District: district_id, A2, A3, A4, A5, A6, A7, A8, A9, A10, A11, A12, A13, A14, A15, A16. ",
     "Loan": "Manages loan-related data, offering insights into each loan's unique identifier, associated account details, approval dates, amounts, durations, and monthly payments. Properties of Loan: loan_id, account_id, date, amount, duration, payments, status. ",
     "Order_": "Manages payment orders, detailing unique identifiers for each order, linked account numbers, and recipient bank details. It captures the bank and account number, the debited amount for each order and categorizes the purpose of each payment. Properties of Order_: order_id, account_id, bank_to, account_to, amount, k_symbol. ",
     "Trans": "Includes transaction management, encompassing details such as transaction identifiers, associated account numbers, and dates of transactions, categorizes transactions, covering a range of activities from insurance payments and statement fees to interest credits, sanctions for negative balances, household payments, pension disbursements, and loan payments; and details about the transaction partner's bank, identified by a unique two-letter code, and their account number. Properties of Trans: trans_id, account_id, date, type, operation, amount, balance, k_symbol, bank, account. "
}
\end{lstlisting}

The list of JSON elements is as follows
\begin{lstlisting}[language=json,firstnumber=1]
['Account', 'Card', 'Client', 'Disp', 'District', 'Loan', 'Order_', 'Trans']
\end{lstlisting}
\end{example}
\begin{example}

Here is an example of a JSON(\{\}, Attributes) used in line~\ref{algo:one:iter:promptA} in Algorithm~\ref{algo:lucy} for the table District to determine relevant attributes (from the  \financial dataset).

\begin{lstlisting}[language=json,firstnumber=1]
{
     "DescriptionField": "Provides a detailed overview of district-level data, essential for regional analysis and decision-making. It includes a unique identifier for each district, along with the district's name and its broader region. The table delves into demographic, economic data and  economic indicators, records crime statistics.",
     "district_id": "location of branch",
     "A2": "district_name",
     "A3": "region",
     "A4": "",
     "A5": "municipality < district < region",
     "A6": "municipality < district < region",
     "A7": "municipality < district < region",
     "A8": "municipality < district < region",
     "A9": "  Description: not useful",
     "A10": "ratio of urban inhabitants",
     "A11": "average salary",
     "A12": "unemployment rate 1995",
     "A13": "unemployment rate 1996",
     "A14": "no. of entrepreneurs per 1000 inhabitants",
     "A15": "no. of committed crimes 1995",
     "A16": "no. of committed crimes 1996"
}
\end{lstlisting}

The list of json elements is as follows
\begin{lstlisting}[language=json,firstnumber=1]
[district_id, A2, A3, A4, A5, A6, A7, A8, A9, A10, A11, A12, A13, A14, A15, A16]
\end{lstlisting}
\end{example}
Moreover, if a table is a root table of a pattern, we provide inner tables and their attribute names so that an \llm can determine the relevance of \snow to the user question. 
\begin{example}
Here is an example of the \snow summary rooted at \respool from the \vlarge benchmark.
\begin{lstlisting}[language=json,firstnumber=1]
"ResourcePool": "Resource pools manage VM resources within a hierarchy, ensuring efficient allocation through configurable settings and states. Properties of ResourcePool: namespace, name, owner, summary, config, config, config.changeVersion, config.entity, config.lastModified, config.scaleDescendantsShares, config.cpualloc, config.cpualloc, config.cpualloc.expandableReservation, config.cpualloc.limit_, config.cpualloc.overheadLimit, config.cpualloc.reservation, config.cpualloc.shares, config.cpualloc, config.memalloc, config.memalloc, config.memalloc.expandableReservation, config.memalloc.limit_, config.memalloc.overheadLimit, config.memalloc.reservation, config.memalloc.shares, config.memalloc, config, runtime, runtime, runtime.overallStatus, runtime.sharesScalable, runtime.cpu, runtime.cpu, runtime.cpu.maxUsage, runtime.cpu.overallUsage, runtime.cpu.reservationUsed, runtime.cpu.reservationUsedForVm, runtime.cpu.unreservedForPool, runtime.cpu.unreservedForVm, runtime.cpu, runtime.memory, runtime.memory, runtime.memory.maxUsage, runtime.memory.overallUsage, runtime.memory.reservationUsed, runtime.memory.reservationUsedForVm, runtime.memory.unreservedForPool, runtime.memory.unreservedForVm, runtime.memory, runtime, ResourcePool_id. "
\end{lstlisting}

\end{example}

\subsubsection{Description of the \ipalgo algorithm.}
\begin{algorithm}
\scriptsize
\caption{\ipalgo}\label{alg:diving}
\begin{algorithmic}[1]
\Require $\Question, \tbl $
\Ensure Relevant tables and attributes in a tree-like pattern rooted at $\tbl$
\State $\stack = [\tbl]$
\State $\relta = \{\}$
\While{$\stack$}
\State $r = \stack.pop()$
\State \textcolor{gray}{// check if $r$ is a leaf in a tree-like pattern}
\If {$\text{leaf}(r)$}
\State $\relta', \text{\_} = \promptA(\Question,\{\},r.\attributes)$
\label{algo:one:iter:promptA:leaf}
\Else
\State \textcolor{gray}{// find children of $r$ in a tree-like pattern}
\State {$\text{children\_tables} =\{\tbl | \tbl \in \dbmshort.tables \cap \text{children}(r)\}$} // children(r) returns descendants of $r$ in the pattern.
\State $\relta', T = \promptA(\Question,\text{children\_tables},r)$
\label{algo:one:iter:promptA}
\State $\stack.push(T)$ 
\EndIf%
\State $\relta =  \relta \cup \relta'$
\EndWhile
\end{algorithmic}
\end{algorithm}
\begin{example}[Full version of Example~\ref{exm:onetwo:match} for the question \qtwo]\label{a:exm:onetwo:match}
Consider \(\qtwo\) from Table~\ref{tab:q1q2}.
Figure~\ref{fig:graph} shows \cdd's core tables. 
\llm identifies \(\respool\) as a relevant table in line~\ref{algo:one:promptA}, along with its attribute \(\respool.\name\). Since \(\respool\) is the root table of a \(\snow\) pattern, we begin to explore the pattern tree in a breadth-first order using \ipalgo in line~\ref{algo:one:dive}. See Figure~\ref{fig:schema} for the structure of the the \(\snow\) pattern. 
\(\respool\) has two child nodes, \(\config\) and \(\runtime\), and several attributes. We then query an \(\llm\) and find that both \(\config\) and \(\runtime\) are relevant as well its attribute \(\respool.\name\). Following the breadth-first search order, we next consider \(\config\) which has descendants \(\ccpu\) and \(\cmem\) and a few attributes. We discover that only one of them, \(\ccpu\), is relevant. We then move to the next table in order, \(\runtime\).
It has two descendants \(\rcpu\) and \(\rmem\) and a few attributes.
We discover that only one of them, \(\rcpu\), is relevant.
Next, we identify relevant attributes of \(\ccpu\) in line~\ref{algo:one:iter:promptA:leaf} (Algorithm~\ref{alg:diving}) and find that \ccpu.\overheadlimit is relevant to the user query.
Finally, we identify relevant attributes of \(\rcpu\) in line~\ref{algo:one:iter:promptA:leaf} in (Algorithm~\ref{alg:diving}) and find that \rcpu.\overallusage is relevant to the user query.
\end{example}
\subsection{The \solve phase}
\label{a:ss:solve}
\subsubsection{Summary view.}
\label{a:ss:solve:view}
Consider again the question \qone~ from Example~\ref{exm:onetwo:match}.
The view \View that corresponds to the green path in Figure~\ref{fig:graph} is shown in the listing in Table~\ref{tab:q4}.
We keep the same set of attributes that \retrieve identified.
In addition, we also perform renaming of all attributes, as we can control the length of the aliases (in case they are too long). 
For example,
\client.\name gets an alias $\client\_\name$, \client.\gender gets  $\client\_\gender$, so on.
\begin{table}
    \centering
    \begin{tabular}{l}
      \\
\begin{lstlisting}[
          label=lst:v,
           language=SQL,
            keywordstyle=\small\color{blue}\ttfamily,
           showspaces=false,
           basicstyle=\small\ttfamily,
           commentstyle=\color{gray},
           mathescape=true
        ]
/*--Summary view $\View$ --*/
create view $\View$ as select  
$\client.\id$ as $\client\_\id$, 
$\client.\name$ as $\client\_\name$, 
$\client.\gender$ as $\client\_\gender$,
$\dc.\name$ as $\dc\_\name$,  
$\dc.\id$ as $\dc\_\id$ 
from $\dc$
join $\compres$     on $\dc.\id$ = $\compres.\dcid$
join $\respool$     on $\compres.\id$ = $\respool.\compresid$
join $\restoclient$ on $\respool.\id$ = $\restoclient.\respoolid$
join $\client$    on $\client.\id$ = $\restoclient.\clientid$
\end{lstlisting} \\
\begin{lstlisting}[
          label=lst:q,
           language=SQL,
            keywordstyle=\small\color{blue}\ttfamily,
           showspaces=false,
           basicstyle=\small\ttfamily,
           commentstyle=\color{gray},
           mathescape=true
        ]
/*--Final query $\Query$--*/
select $\client.\name$, 
$\dc.\name$
from $\View$ where  $\dc.\name$ like 'dev%';
\end{lstlisting}
    \end{tabular}
 \caption{$\solve$ and $\tvs$ results for $\qone$.}
 \label{tab:q4}
\end{table}
\subsection{The \tvs phase}
\subsubsection{\promptC}
\label{a:ss:promptC}

Here is \promptC that we use in the final phase \tvs (Algorithm~\ref{algo:lucy}, line~\ref{algo:three:prompt}).  The function name() returns name of the view \View.
\begin{tcolorbox}[colback=white]
\paragraph{Inputs:}  Question, \View
\paragraph{\promptC}
I created a view table <name(\View)> with all relevant information. \
Here is a view <\View>. \
Please write MySQL query to  name(\View) view to answer the following question: <Question>. \
Use only name(\View) columns in the query. \
Absolutely NO columns renaming. \
Absolutely NO HAVING operators. \
Absolutely NO COUNT(*). \
Output query that I can run via python interface. Output '```sql...'. Do not explain.
\paragraph{Returns:} SQL
\end{tcolorbox}

We used a few  assertive statements that we discuss next. 'Absolutely NO column renaming' means that we want to use aliases in the view table to form a valid SQL query.
The statement 'Absolutely NO HAVING operators.' reflects our observation that \(\gptfourT\) cannot generate valid SQL when using HAVING in combination with GROUP BY. It is a subject of future research to deal with MySQL constraints, so we encourage \(\tvs\) to avoid this operator.
Finally, we discourage the use of COUNT(*), `Absolutely NO COUNT(*)', to ensure that \gptfourT
focuses on counting the entities specified in the user's question.

We noticed that better results are obtained if we provide a description of tables that are used to generate this view together with their relevant attributes. Here is an extended version of \promptC where we provide relevant tables and their attributes that are used to obtain the \View. We also provide an evidence if available.
\begin{tcolorbox}[colback=white]
\paragraph{Inputs:}  Question, \View, DB\_SCHEMA
\paragraph{\promptC' (with evidence and a part of the schema)}
Here is a SQL schema for in MySQL: <DB\_SCHEMA>
I created a view table <name(\View)> with all relevant information. \
Here is a view <\View>. \
Please write MySQL query to name(\View) view to answer the following question: <Question>. \
Additional knowledge to answer:  <Evidence>  \
Use only name(\View) columns in the query. \
Absolutely NO columns renaming. \
Absolutely NO HAVING operators. \
Absolutely NO COUNT(*). \
Output query that I can run via python interface. Output '```sql...'. Do not explain.
\paragraph{Returns:} SQL
\end{tcolorbox}

\section{Experimental evaluation (additional materials)}
\subsection{Setup}
\label{a:exp:setup}
We run experiments on a laptop Intel(r) Core 2.40Hz and 32GB of memory.
For \nsql we use the largest model with 7B parameters (NumbersStation/nsql-llama-2-7B~\citep{nsql}).
For \gptfourT and \sys, we use the `gpt-4-0125-preview' model as a \llm and set the temperature to 0.2 . We do not fine-tune a \llm.  We require 20 answers from \gptfourT for each question. If the number of correct answers is more than 5, then we count that benchmark as solved. 

In the case of \sys, we require 5 answers for each \gpt call for the \retrieve phase. We sort tables based on the number of occurrences in these answers and take at most 8 candidates among relevant tables from each \promptA output. Similarly to  \gptfourT, we require 20 answers from \tvs and decide on the success as described above. We use ORTools as a constraint solver~\citep{cpsatlp}.

We support MySQL as a relational database. However, BIRD uses SQLite. We automatically, converted queries from sqlite to MySQL.

We provide all benchmarks and their results in the supplementary materials.
\subsection{\insurance}
\label{a:ss:insurance}
\paragraph{Note on the database.}
There are a few issues with broken relational constraints due to missing tables, as reported~\citep{insuranceissues}, which we fixed by adding the missing tables from the original database.

\paragraph{Extended schema examples.}
Example of tables extended with comments that describe each attribute for the \insurance benchmark.

\begin{lstlisting}[
          label=a:lst:ex1,
           language=SQL,
           showspaces=false,
            keywordstyle=\small\color{blue}\ttfamily,
           basicstyle=\small\ttfamily,
           commentstyle=\color{gray},
        mathescape=true
        ]
CREATE TABLE Claim_Amount
( 
	Claim_Amount_Identifier bigint  NOT NULL COMMENT Claim Amount Identifier is the unique identifier of the financial amount reserved, paid, or collected in connection with a claim. The money being paid or collected for settling a claim and paying the claimants, reinsurers, other insurers, and other interested parties. Claim amounts are classified by various attributes.,
	Claim_Identifier     int  NOT NULL COMMENT Claim Identifier is the unique identifier for a Claim.,
	Claim_Offer_Identifier int  NULL COMMENT Claim Offer Identifier is the unique identifier for a Claim Offer.,
	Amount_Type_Code     varchar(20)  NULL COMMENT Amount Type Code defines the category to which a monetary amount will be applied. Example:  premium, commission, tax, surcharge.,
	Event_Date           datetime  NULL COMMENT Event Date is the date on which a transaction or insurance-related happening takes place.,
	Claim_Amount         decimal(15,2)  NULL COMMENT The money being paid or collected for settling a claim and paying the claimants, reinsurers, other insurers, and other interested parties. Claim amounts are classified by various attributes.,
	Insurance_Type_Code  char(1)  NULL COMMENT  Insurance Type Code represents the category under which risk is assumed.  Examples: Direct for policies directly issued by a company; Assumed for risks assumed from another company; Ceded for portions of risk ceded to another insurer.,
	 PRIMARY KEY (Claim_Amount_Identifier ASC),
	 FOREIGN KEY (Claim_Offer_Identifier) REFERENCES Claim_Offer(Claim_Offer_Identifier),
 FOREIGN KEY (Claim_Identifier) REFERENCES Claim(Claim_Identifier)
)
\end{lstlisting} 
\begin{lstlisting}[
          label=a:lst:ex2,
           language=SQL,
           showspaces=false,
            keywordstyle=\small\color{blue}\ttfamily,
           basicstyle=\small\ttfamily,
           commentstyle=\color{gray},
        mathescape=true
        ]

CREATE TABLE Claim_Reserve
( 
	Claim_Amount_Identifier bigint  NOT NULL COMMENT Claim Amount Identifier is the unique identifier of the financial amount reserved, paid, or collected in connection with a claim. The amount of expected loss over the life of the Claim.,
	 PRIMARY KEY (Claim_Amount_Identifier ASC),
	 FOREIGN KEY (Claim_Amount_Identifier) REFERENCES Claim_Amount(Claim_Amount_Identifier)
)
\end{lstlisting} 
\subsubsection{Challenging questions}
\label{a:sss:insurance}
In this section, we present 13 questions that \sys found challenging to answer and identify reasons for these failures. 
\begin{tcolorbox}
\paragraph{Question1:} What are the loss payment, loss reserve, expense payment, expense reserve amount by claim number and corresponding policy number, policy holder, premium amount paid, the catastrophe it had, and the agent who sold it?

\paragraph{Reason:} \ml "policy holder" and "agent" require a look up to the same table Agreement\_Party\_Role.
\end{tcolorbox}

\begin{tcolorbox}
\paragraph{Question2:} What are the total loss, which is the sum of loss payment, loss reserve, expense payment, expense reserve amount by claim number and corresponding policy number, policy holder and premium amount paid?

\paragraph{Reason:} \pone Phase 1 misses the relevant table Agreement\_Party\_Role.
\end{tcolorbox}

\begin{tcolorbox}
\paragraph{Question3:} What is the total amount of premiums that a policy holder has paid?

\paragraph{Reason:} \pthree Phase 3 makes a mistake in  the \groupby clause.
\end{tcolorbox}

\begin{tcolorbox}
\paragraph{Question4:} What are the total loss, which is the sum of loss payment, loss reserve, expense payment, expense reserve amount by catastrophe and policy number?

\paragraph{Reason:}  \aq
By "by catastrophe", the user means to output Catastrophe's attribute Name. However, Phase 1 identifies  Catastrophe's attribute Identifier as relevant instead of Name.
\end{tcolorbox}

\begin{tcolorbox}
\paragraph{Question5:} What is the average policy size which is the the total amount of premium divided by the number of policies?

\paragraph{Reason:}  \aq
The definition of average is not standard, as the same policy can have multiple \amount values.
\end{tcolorbox}

\begin{tcolorbox}
\paragraph{Question6:} What are the loss payment, loss reserve, expense payment, expense reserve amount by claim number and corresponding policy number, policy holder, premium amount paid and the agent who sold it?

\paragraph{Reason:}  \ml
\end{tcolorbox}

\begin{tcolorbox}
\paragraph{Question7:} Return agents and the policy they have sold that have had a claim and the corresponding catastrophe it had.

\paragraph{Reason:}  \aq
The output includes Company\_Claim\_Number, although this information is not specified in the question.
\end{tcolorbox}

\begin{tcolorbox}
\paragraph{Question8:} What is the loss ratio of each policy and agent who sold it by policy number and agent id?

\paragraph{Reason:} \aq
"the loss ratio" is a complex formula here, making it hard to guess without its proper specification.
\end{tcolorbox}

\begin{tcolorbox}
\paragraph{Question9:} What are all the premiums that have been paid by policy holders?

\paragraph{Reason:}   \aq
Policy.Policy\_Number and Party\_Identifier should be included in the output. But they are not specified in the question.
\end{tcolorbox}

\begin{tcolorbox}
\paragraph{Question10:} What are the loss payment, loss reserve, expense payment, expense reserve amount by claim number and corresponding policy number, policy holder and premium amount paid?

\paragraph{Reason:} \pone Phase 1 misses the relevant table Agreement\_Party\_Role.
\end{tcolorbox}

\begin{tcolorbox}
\paragraph{Question11:}  What is the loss ratio, number of claims, total loss by policy number and premium where total loss is the sum of loss payment, loss reserve, expense payment, expense reserve amount and loss ratio is total loss divided by premium?

\paragraph{Reason:} \pone Phase 1 misses the relevant table Policy. 
\end{tcolorbox}

\begin{tcolorbox}
\paragraph{Question12:} What are the total loss, which is the sum of loss payment, loss reserve, expense payment, expense reserve amount by claim number, catastrophe and corresponding policy number?

\paragraph{Reason:}  \pone Phase 1 misses the relevant table Catastrophe. 
\end{tcolorbox}

\begin{tcolorbox}
\paragraph{Question13:} What is the total amount of premiums that a policy holder has paid by policy number?

\paragraph{Reason:}  \aq
Party\_Identifier is included in the output. But it is not specified in the question.
\end{tcolorbox}

\subsection{BIRD datasets}
\label{a:ss:bird}

\subsubsection{Additional notes on the dataset.}
\paragraph{Note on \dbm.}
We used attribute descriptions available in BIRD in \dbm. We also build table descriptions in the following way. We provided the description from BIRD to an \(\llm\) to generate a short summary description using \promptD defined in Section~\ref{a:ss:promptD}.

\paragraph{Note on datasets.} It has been shown that there are a number of incorrect ground truth SQLs in BIRD datasets~\citep{birdissues, wretblad2024understanding}. For example, \cite{wretblad2024understanding} found that 72 out of 106 benchmark questions in \financial have errors of various types. Most of the issues have been reported to the authors from multiple sources, and we also reported additional problems via private communication. The authors acknowledge these issues and are working on them. To provide an example we reported from \formula:
\begin{itemize}
\item Question: `Where can the introduction of the races held on Circuit de Barcelona-Catalunya be found?'
\item Ground truth SQL: \select~~\distinct \textcolor{red}{circuits.url} \from circuits \ijn races \on races.circuitId = circuits.circuitId \where circuits.name = 'Circuit de Barcelona-Catalunya'.
\item The issue is that \select should be on \textcolor{red}{race.url} rather than \textcolor{blue}{circuits.url} as the user requests information about the race, not the circuit.
\end{itemize}

On top of that, there are \emph{logical inconsistencies} in ground truth answers for the \financial  dataset. Often, users ask for information about clients' accounts. Client and account tables have a \manymany relationship modeled using an additional table disp. At the same time, they are both related to a lookup table district. Unfortunately, many ground truth SQL statements perform a join between clients and accounts via the district table, which is incorrect. Let's consider an example.
\begin{itemize}
\item Question: `Please provide the IDs of the 3 female clients with the largest loans.'
\item Ground truth SQL: 
\select T1.client\_id 
\from client \as T1 
\ijn account \as T2 \on \textcolor{red}{T1.district\_id = T2.district\_id}
\ijn loan \as T3 \on T2.account\_id = T3.account\_id 
\where T1.gender = 'F' 
\orderby  T3.amount DESC LIMIT 3
\item The issue is that the answer relates clients and accounts that have the same \textcolor{red}{district\_id}. However, this does not mean that the client is related to the account.
\end{itemize}
As the authors are working on corrections, we analyzed the reported issues and manually corrected the ground truth. We only adjusted the SQL ground truth values to match the user questions; we did not alter the questions or evidences. We provide the corrected benchmarks in the supplementary material. Specifically, we corrected the \financial and \formula benchmarks and used the correct answers to evaluate all methods. Interestingly, the performance of all frameworks improved by a few percentage points when we fixed these ground truth SQL statements.
\subsubsection{\promptD}
\label{a:ss:promptD}
Here is \promptD that we use to generate tables summaries for \financial and \formula datasets.
\begin{tcolorbox}
\paragraph{\promptD} 
Give me a very brief description of the <NAME> table.
\end{tcolorbox}

\begin{tcolorbox}
\paragraph{Example for the Loan table (the \financial database)} 

\paragraph{Prompt:} Give me a very brief description of the Loan table.

original\_column\_name,column\_name,column\_description,data\_format,
   value\_description

loan\_id,,the id number identifying the loan data,integer,

account\_id,,the id number identifying the account,integer,

date,,the date when the loan is approved,date,

amount,,approved amount,integer,unit: US dollar

duration,,loan duration,integer,unit: month

payments,monthly payments,monthly payments,real,unit: month

status,,repayment status,text,"'A' stands for contract finished, no problems; 
'B' stands for contract finished, loan not paid; 'C' stands for running contract, OK so far; 'D' stands for running contract, client in debt"

\paragraph{Return:} The Loan table manages loan-related data, offering insights into each loan's unique identifier, associated account details, approval dates, amounts, durations, and monthly payments.
\end{tcolorbox}

\subsubsection{Challenging questions}
\label{a:sss:financial}
We discuss three major groups of challenging  questions with examples.

The first group contains ambiguous questions.
Here are a few examples.
\begin{tcolorbox}
\paragraph{Question:} List out the no. of districts that have female average salary is more than 6000 but less than 10000?

\paragraph{Reason:} \aq
`no. of districts' refers to the district number based on the ground truth. However, \(\sys\) counts the number of districts.
\end{tcolorbox}

\begin{tcolorbox}
\paragraph{Question:} W that the client whose card was opened in 1996/10/21 made?

\paragraph{Reason:} \aq 
\(\sys\) filters on `card issued date', while ground truth filters on `account opened date'. However, the user is indeed asking about `card open date' in this question. This issue was also independently observed in~\citep{noise}. 
\end{tcolorbox}

The second group contains complex filtering, ordering, and/or formulas to compute. 
Here are a few examples.
\begin{tcolorbox}
\paragraph{Question:} List out the account numbers of clients who are youngest and have highest average salary?

\paragraph{Reason:} \pthree
There are two filtering conditions that have to be applied in order. First, we find the youngest clients, then select the one with the highest average salary among them. \(\sys\) treats these conditions as a conjunction, resulting in an empty output.
\end{tcolorbox}

\begin{tcolorbox}
\paragraph{Question:} List out the account numbers of female clients who are oldest and has lowest average salary, calculate the gap between this lowest average salary with the highest average salary?

\paragraph{Reason:} \pthree 
Two filtering conditions are required: first, in descending order, and then in ascending order. However, \(\sys\) fails to perform them in this sequence.
\end{tcolorbox}

\begin{tcolorbox}
\paragraph{Question:} For the client who applied the biggest loan, what was his/her first amount of transaction after opened the account.

\paragraph{Reason:} \pthree 
Two filtering conditions are required: first, in ascending order, and then in descending order. However, \(\sys\) fails to perform them in this sequence.
\end{tcolorbox}

The third group contains questions where the \retrieve phase either adds an extra table, or occasionally misses a table or attributes.
Here is an example.
\begin{tcolorbox}
\paragraph{Question:} How many accounts have an owner disposition and request for a statement to be generated upon a transaction?

\paragraph{Reason:} \pone 
\(\sys\) identifies "Tran" (transaction) as a relevant table, but it is not needed to answer the query.
\end{tcolorbox}

\subsection{Cloud resources}
\paragraph{Note on the cost of running.} One note here is that \gpt and \chatfour models are costly to run. For example, in the \vlarge experiment, the costs are as follows: \chatfour costs \$15, \gptfourT \$2, and \gptfourTS \$5, while \sys costs \$0.5. 

\newpage
\section*{NeurIPS Paper Checklist}

\begin{enumerate}

\item {\bf Claims}
    \item[] Question: Do the main claims made in the abstract and introduction accurately reflect the paper's contributions and scope?
    \item[] Answer:  \answerYes{} 
    \item[] Justification: Yes, we do.
    \item[] Guidelines:
    \begin{itemize}
        \item The answer NA means that the abstract and introduction do not include the claims made in the paper.
        \item The abstract and/or introduction should clearly state the claims made, including the contributions made in the paper and important assumptions and limitations. A No or NA answer to this question will not be perceived well by the reviewers. 
        \item The claims made should match theoretical and experimental results, and reflect how much the results can be expected to generalize to other settings. 
        \item It is fine to include aspirational goals as motivation as long as it is clear that these goals are not attained by the paper. 
    \end{itemize}

\item {\bf Limitations}
    \item[] Question: Does the paper discuss the limitations of the work performed by the authors?
    \item[] Answer:  \answerYes{} 
    \item[] Justification: Yes, see Section~\ref{ss:discussion}
    \item[] Guidelines:
    \begin{itemize}
        \item The answer NA means that the paper has no limitation while the answer No means that the paper has limitations, but those are not discussed in the paper. 
        \item The authors are encouraged to create a separate "Limitations" section in their paper.
        \item The paper should point out any strong assumptions and how robust the results are to violations of these assumptions (e.g., independence assumptions, noiseless settings, model well-specification, asymptotic approximations only holding locally). The authors should reflect on how these assumptions might be violated in practice and what the implications would be.
        \item The authors should reflect on the scope of the claims made, e.g., if the approach was only tested on a few datasets or with a few runs. In general, empirical results often depend on implicit assumptions, which should be articulated.
        \item The authors should reflect on the factors that influence the performance of the approach. For example, a facial recognition algorithm may perform poorly when image resolution is low or images are taken in low lighting. Or a speech-to-text system might not be used reliably to provide closed captions for online lectures because it fails to handle technical jargon.
        \item The authors should discuss the computational efficiency of the proposed algorithms and how they scale with dataset size.
        \item If applicable, the authors should discuss possible limitations of their approach to address problems of privacy and fairness.
        \item While the authors might fear that complete honesty about limitations might be used by reviewers as grounds for rejection, a worse outcome might be that reviewers discover limitations that aren't acknowledged in the paper. The authors should use their best judgment and recognize that individual actions in favor of transparency play an important role in developing norms that preserve the integrity of the community. Reviewers will be specifically instructed to not penalize honesty concerning limitations.
    \end{itemize}

\item {\bf Theory Assumptions and Proofs}
    \item[] Question: For each theoretical result, does the paper provide the full set of assumptions and a complete (and correct) proof?
    \item[] Answer: \answerYes{} 
    \item[] Justification: We model a part of the problem as an optimization problem and provide formal encoding. See Section~\ref{ss:solve}.
    \item[] Guidelines:
    \begin{itemize}
        \item The answer NA means that the paper does not include theoretical results. 
        \item All the theorems, formulas, and proofs in the paper should be numbered and cross-referenced.
        \item All assumptions should be clearly stated or referenced in the statement of any theorems.
        \item The proofs can either appear in the main paper or the supplemental material, but if they appear in the supplemental material, the authors are encouraged to provide a short proof sketch to provide intuition. 
        \item Inversely, any informal proof provided in the core of the paper should be complemented by formal proofs provided in appendix or supplemental material.
        \item Theorems and Lemmas that the proof relies upon should be properly referenced. 
    \end{itemize}

    \item {\bf Experimental Result Reproducibility}
    \item[] Question: Does the paper fully disclose all the information needed to reproduce the main experimental results of the paper to the extent that it affects the main claims and/or conclusions of the paper (regardless of whether the code and data are provided or not)?
    \item[] Answer: \answerYes{} 
    \item[] Justification: Yes, we describe all algorithms and an optimization model.
    \item[] Guidelines:
    \begin{itemize}
        \item The answer NA means that the paper does not include experiments.
        \item If the paper includes experiments, a No answer to this question will not be perceived well by the reviewers: Making the paper reproducible is important, regardless of whether the code and data are provided or not.
        \item If the contribution is a dataset and/or model, the authors should describe the steps taken to make their results reproducible or verifiable. 
        \item Depending on the contribution, reproducibility can be accomplished in various ways. For example, if the contribution is a novel architecture, describing the architecture fully might suffice, or if the contribution is a specific model and empirical evaluation, it may be necessary to either make it possible for others to replicate the model with the same dataset, or provide access to the model. In general. releasing code and data is often one good way to accomplish this, but reproducibility can also be provided via detailed instructions for how to replicate the results, access to a hosted model (e.g., in the case of a large language model), releasing of a model checkpoint, or other means that are appropriate to the research performed.
        \item While NeurIPS does not require releasing code, the conference does require all submissions to provide some reasonable avenue for reproducibility, which may depend on the nature of the contribution. For example
        \begin{enumerate}
            \item If the contribution is primarily a new algorithm, the paper should make it clear how to reproduce that algorithm.
            \item If the contribution is primarily a new model architecture, the paper should describe the architecture clearly and fully.
            \item If the contribution is a new model (e.g., a large language model), then there should either be a way to access this model for reproducing the results or a way to reproduce the model (e.g., with an open-source dataset or instructions for how to construct the dataset).
            \item We recognize that reproducibility may be tricky in some cases, in which case authors are welcome to describe the particular way they provide for reproducibility. In the case of closed-source models, it may be that access to the model is limited in some way (e.g., to registered users), but it should be possible for other researchers to have some path to reproducing or verifying the results.
        \end{enumerate}
    \end{itemize}

\item {\bf Open access to data and code}
    \item[] Question: Does the paper provide open access to the data and code, with sufficient instructions to faithfully reproduce the main experimental results, as described in supplemental material?
    \item[] Answer:  \answerYes{}
    \item[] Justification: We provide the data in supplementary materials and describe prompts. We will make code publicly available.
    \item[] Guidelines:
    \begin{itemize}
        \item The answer NA means that paper does not include experiments requiring code.
        \item Please see the NeurIPS code and data submission guidelines (\url{https://nips.cc/public/guides/CodeSubmissionPolicy}) for more details.
        \item While we encourage the release of code and data, we understand that this might not be possible, so “No” is an acceptable answer. Papers cannot be rejected simply for not including code, unless this is central to the contribution (e.g., for a new open-source benchmark).
        \item The instructions should contain the exact command and environment needed to run to reproduce the results. See the NeurIPS code and data submission guidelines (\url{https://nips.cc/public/guides/CodeSubmissionPolicy}) for more details.
        \item The authors should provide instructions on data access and preparation, including how to access the raw data, preprocessed data, intermediate data, and generated data, etc.
        \item The authors should provide scripts to reproduce all experimental results for the new proposed method and baselines. If only a subset of experiments are reproducible, they should state which ones are omitted from the script and why.
        \item At submission time, to preserve anonymity, the authors should release anonymized versions (if applicable).
        \item Providing as much information as possible in supplemental material (appended to the paper) is recommended, but including URLs to data and code is permitted.
    \end{itemize}

\item {\bf Experimental Setting/Details}
    \item[] Question: Does the paper specify all the training and test details (e.g., data splits, hyperparameters, how they were chosen, type of optimizer, etc.) necessary to understand the results?
    \item[] Answer: \answerYes{} 
    \item[] Justification: We specified parameters of prompts. We do not train new models.
    \item[] Guidelines:
    \begin{itemize}
        \item The answer NA means that the paper does not include experiments.
        \item The experimental setting should be presented in the core of the paper to a level of detail that is necessary to appreciate the results and make sense of them.
        \item The full details can be provided either with the code, in appendix, or as supplemental material.
    \end{itemize}

\item {\bf Experiment Statistical Significance}
    \item[] Question: Does the paper report error bars suitably and correctly defined or other appropriate information about the statistical significance of the experiments?
    \item[] Answer:  \answerNA{} 
    \item[] Justification: We provide details for \sys and \gptfourT. Existing methods either provide their results as a single answer~\citep{birdleader} or are too costly to run multiple times.
    \item[] Guidelines:
    \begin{itemize}
        \item The answer NA means that the paper does not include experiments.
        \item The authors should answer "Yes" if the results are accompanied by error bars, confidence intervals, or statistical significance tests, at least for the experiments that support the main claims of the paper.
        \item The factors of variability that the error bars are capturing should be clearly stated (for example, train/test split, initialization, random drawing of some parameter, or overall run with given experimental conditions).
        \item The method for calculating the error bars should be explained (closed form formula, call to a library function, bootstrap, etc.)
        \item The assumptions made should be given (e.g., Normally distributed errors).
        \item It should be clear whether the error bar is the standard deviation or the standard error of the mean.
        \item It is OK to report 1-sigma error bars, but one should state it. The authors should preferably report a 2-sigma error bar than state that they have a 96\% CI, if the hypothesis of Normality of errors is not verified.
        \item For asymmetric distributions, the authors should be careful not to show in tables or figures symmetric error bars that would yield results that are out of range (e.g. negative error rates).
        \item If error bars are reported in tables or plots, The authors should explain in the text how they were calculated and reference the corresponding figures or tables in the text.
    \end{itemize}

\item {\bf Experiments Compute Resources}
    \item[] Question: For each experiment, does the paper provide sufficient information on the computer resources (type of compute workers, memory, time of execution) needed to reproduce the experiments?
    \item[] Answer: \answerYes{} 
    \item[] Justification: Yes, we describe the experimental setup.
    \item[] Guidelines:
    \begin{itemize}
        \item The answer NA means that the paper does not include experiments.
        \item The paper should indicate the type of compute workers CPU or GPU, internal cluster, or cloud provider, including relevant memory and storage.
        \item The paper should provide the amount of compute required for each of the individual experimental runs as well as estimate the total compute. 
        \item The paper should disclose whether the full research project required more compute than the experiments reported in the paper (e.g., preliminary or failed experiments that didn't make it into the paper). 
    \end{itemize}
    
\item {\bf Code Of Ethics}
    \item[] Question: Does the research conducted in the paper conform, in every respect, with the NeurIPS Code of Ethics \url{https://neurips.cc/public/EthicsGuidelines}?
    \item[] Answer: \answerYes{} 
    \item[] Justification: 
    \item[] Guidelines:
    \begin{itemize}
        \item The answer NA means that the authors have not reviewed the NeurIPS Code of Ethics.
        \item If the authors answer No, they should explain the special circumstances that require a deviation from the Code of Ethics.
        \item The authors should make sure to preserve anonymity (e.g., if there is a special consideration due to laws or regulations in their jurisdiction).
    \end{itemize}

\item {\bf Broader Impacts}
    \item[] Question: Does the paper discuss both potential positive societal impacts and negative societal impacts of the work performed?
    \item[] Answer:  \answerYes{} 
    \item[] Justification: We believe it has a positive impact as we enhance users with new capabilities.
    \item[] Guidelines:
    \begin{itemize}
        \item The answer NA means that there is no societal impact of the work performed.
        \item If the authors answer NA or No, they should explain why their work has no societal impact or why the paper does not address societal impact.
        \item Examples of negative societal impacts include potential malicious or unintended uses (e.g., disinformation, generating fake profiles, surveillance), fairness considerations (e.g., deployment of technologies that could make decisions that unfairly impact specific groups), privacy considerations, and security considerations.
        \item The conference expects that many papers will be foundational research and not tied to particular applications, let alone deployments. However, if there is a direct path to any negative applications, the authors should point it out. For example, it is legitimate to point out that an improvement in the quality of generative models could be used to generate deepfakes for disinformation. On the other hand, it is not needed to point out that a generic algorithm for optimizing neural networks could enable people to train models that generate Deepfakes faster.
        \item The authors should consider possible harms that could arise when the technology is being used as intended and functioning correctly, harms that could arise when the technology is being used as intended but gives incorrect results, and harms following from (intentional or unintentional) misuse of the technology.
        \item If there are negative societal impacts, the authors could also discuss possible mitigation strategies (e.g., gated release of models, providing defenses in addition to attacks, mechanisms for monitoring misuse, mechanisms to monitor how a system learns from feedback over time, improving the efficiency and accessibility of ML).
    \end{itemize}
    
\item {\bf Safeguards}
    \item[] Question: Does the paper describe safeguards that have been put in place for responsible release of data or models that have a high risk for misuse (e.g., pretrained language models, image generators, or scraped datasets)?
    \item[] Answer:  \answerNA{}.
    \item[] Justification: 
    \item[] Guidelines:
    \begin{itemize}
        \item The answer NA means that the paper poses no such risks.
        \item Released models that have a high risk for misuse or dual-use should be released with necessary safeguards to allow for controlled use of the model, for example by requiring that users adhere to usage guidelines or restrictions to access the model or implementing safety filters. 
        \item Datasets that have been scraped from the Internet could pose safety risks. The authors should describe how they avoided releasing unsafe images.
        \item We recognize that providing effective safeguards is challenging, and many papers do not require this, but we encourage authors to take this into account and make a best faith effort.
    \end{itemize}

\item {\bf Licenses for existing assets}
    \item[] Question: Are the creators or original owners of assets (e.g., code, data, models), used in the paper, properly credited and are the license and terms of use explicitly mentioned and properly respected?
    \item[] Answer:  \answerYes{} 
    \item[] Justification: 
    \item[] Guidelines:
    \begin{itemize}
        \item The answer NA means that the paper does not use existing assets.
        \item The authors should cite the original paper that produced the code package or dataset.
        \item The authors should state which version of the asset is used and, if possible, include a URL.
        \item The name of the license (e.g., CC-BY 4.0) should be included for each asset.
        \item For scraped data from a particular source (e.g., website), the copyright and terms of service of that source should be provided.
        \item If assets are released, the license, copyright information, and terms of use in the package should be provided. For popular datasets, \url{paperswithcode.com/datasets} has curated licenses for some datasets. Their licensing guide can help determine the license of a dataset.
        \item For existing datasets that are re-packaged, both the original license and the license of the derived asset (if it has changed) should be provided.
        \item If this information is not available online, the authors are encouraged to reach out to the asset's creators.
    \end{itemize}

\item {\bf New Assets}
    \item[] Question: Are new assets introduced in the paper well documented and is the documentation provided alongside the assets?
    \item[] Answer: \answerYes{} 
    \item[] Justification: We provide all benchmarks in supplementary materials.
    \item[] Guidelines:
    \begin{itemize}
        \item The answer NA means that the paper does not release new assets.
        \item Researchers should communicate the details of the dataset/code/model as part of their submissions via structured templates. This includes details about training, license, limitations, etc. 
        \item The paper should discuss whether and how consent was obtained from people whose asset is used.
        \item At submission time, remember to anonymize your assets (if applicable). You can either create an anonymized URL or include an anonymized zip file.
    \end{itemize}

\item {\bf Crowdsourcing and Research with Human Subjects}
    \item[] Question: For crowdsourcing experiments and research with human subjects, does the paper include the full text of instructions given to participants and screenshots, if applicable, as well as details about compensation (if any)? 
    \item[] Answer:  \answerNA{} 
    \item[] Justification:
    \item[] Guidelines:
    \begin{itemize}
        \item The answer NA means that the paper does not involve crowdsourcing nor research with human subjects.
        \item Including this information in the supplemental material is fine, but if the main contribution of the paper involves human subjects, then as much detail as possible should be included in the main paper. 
        \item According to the NeurIPS Code of Ethics, workers involved in data collection, curation, or other labor should be paid at least the minimum wage in the country of the data collector. 
    \end{itemize}

\item {\bf Institutional Review Board (IRB) Approvals or Equivalent for Research with Human Subjects}
    \item[] Question: Does the paper describe potential risks incurred by study participants, whether such risks were disclosed to the subjects, and whether Institutional Review Board (IRB) approvals (or an equivalent approval/review based on the requirements of your country or institution) were obtained?
    \item[] Answer:  \answerNA{} 
    \item[] Justification: 
    \item[] Guidelines:
    \begin{itemize}
        \item The answer NA means that the paper does not involve crowdsourcing nor research with human subjects.
        \item Depending on the country in which research is conducted, IRB approval (or equivalent) may be required for any human subjects research. If you obtained IRB approval, you should clearly state this in the paper. 
        \item We recognize that the procedures for this may vary significantly between institutions and locations, and we expect authors to adhere to the NeurIPS Code of Ethics and the guidelines for their institution. 
        \item For initial submissions, do not include any information that would break anonymity (if applicable), such as the institution conducting the review.
    \end{itemize}

\end{enumerate}

\end{document}